\documentclass{article}

\usepackage{microtype}
\usepackage{graphicx}
\usepackage{subfigure}
\usepackage{booktabs} %
\usepackage{multirow} 
\usepackage{hyperref}

\usepackage[accepted]{icml2024}

\usepackage{amsmath}
\usepackage{amssymb}
\usepackage{mathtools}
\usepackage{amsthm}

\usepackage[capitalize,noabbrev]{cleveref}

\theoremstyle{plain}

\theoremstyle{definition}

\theoremstyle{remark}

\usepackage[textsize=tiny]{todonotes}

\usepackage{nicefrac}       %
\usepackage{microtype}      %
\PassOptionsToClass{usenames,dvipsnames}{xcolor}
\definecolor{TUblue}{RGB}{0,105,170}

\usepackage{comment}
\newcommand{\wu}[1]{{\color{red} [\textbf{Wu:} #1]}}

\usetikzlibrary{calc,arrows.meta}
\definecolor{VectorBlue}{RGB}{59,69,227}
\colorlet{maincolor}{VectorBlue}
\makeatletter%
\ifcsname c@listlen\endcsname%
\else \newcounter{listlen}\fi%
\makeatother%
\def\setListlen#1{%
  \setcounter{listlen}{0}%
  \foreach \x in #1{%
    \stepcounter{listlen}%
  }%
}%
\tikzset{%
  tensor/.style={%
    rectangle,%
    minimum width=4ex,%
    minimum height=3.5ex,%
    inner sep=0.3ex,%
    rounded corners,%
    fill=maincolor!30!white,%
    draw=black, thick%
  },%
  index/.style={%
    fill=white,%
    inner sep=0.3ex%
  },%
}%
\def\tensornode#1#2#3#4#5#6{%
  \node[tensor] (#1) {#2};%
  \setListlen{#3}%
  \foreach[count=\n] \name in #3%
  \draw ($ (#1.north east)!{\n/(\the\value{listlen}+1)}!(#1.south east) $) to
  ++(1.5ex, 0) coordinate (#1-\name);%
  \setListlen{#4}%
  \foreach[count=\n] \name in #4%
  \draw ($ (#1.north west)!{\n/(\the\value{listlen}+1)}!(#1.north east) $) to
  ++(0, 1.5ex) coordinate (#1-\name);%
  \setListlen{#5}%
  \foreach[count=\n] \name in #5%
  \draw ($ (#1.north west)!{\n/(\the\value{listlen}+1)}!(#1.south west) $) to
  ++(-1.5ex, 0) coordinate (#1-\name);%
  \setListlen{#6}%
  \foreach[count=\n] \name in #6%
  \draw ($ (#1.south west)!{\n/(\the\value{listlen}+1)}!(#1.south east) $) to
  ++(0, -1.5ex) coordinate (#1-\name);%
}%
\def\W{%
  \tensornode{W}%
  {$\tW$}%
  {{cout}}%
  {{k2, k1}}%
  {{cin}}%
  {{}}%
}%
\def\contract#1#2#3#4#5{%
  (#1-#3) edge[#5] node[index] {#4} (#2-#3)%
}%

\newcommand{\stepsize}{\beta}

\newcommand\cut[1]{}

\newcommand{\squishlist}{
   \begin{list}{$\bullet$}
    { \setlength{\itemsep}{0pt}      \setlength{\parsep}{3pt}
      \setlength{\topsep}{3pt}       \setlength{\partopsep}{0pt}
      \setlength{\leftmargin}{1.5em} \setlength{\labelwidth}{1em}
      \setlength{\labelsep}{0.5em} } }

\newcommand{\squishlisttwo}{
   \begin{list}{$\bullet$}
    { \setlength{\itemsep}{0pt}    \setlength{\parsep}{0pt}
      \setlength{\topsep}{0pt}     \setlength{\partopsep}{0pt}
      \setlength{\leftmargin}{2em} \setlength{\labelwidth}{1.5em}
      \setlength{\labelsep}{0.5em} } }

\newcommand{\squishend}{
    \end{list}  }

\newtheorem{claim}{Claim}{}
{}
{}
{}
\newtheorem{defn}{Definition}{}

\newcommand{\half}{\mbox{$\frac{1}{2}$}}

\newcommand{\real}{\mbox{$\mathbb{R}$}}

\newcommand{\myvec}[1]{\mbox{$\mathbf{#1}$}}
\newcommand{\myvecsym}[1]{\mbox{$\boldsymbol{#1}$}}

\newcommand{\veta}{\mbox{$\myvecsym{\eta}$}}

\newcommand{\vmu}{\mbox{$\myvecsym{\mu}$}}

\newcommand{\vtheta}{\mbox{$\myvecsym{\theta}$}}

\newcommand{\vtau}{\mbox{$\myvecsym{\tau}$}}

\newcommand{\va}{\mbox{$\myvec{a}$}}

\newcommand{\vc}{\mbox{$\myvec{c}$}}

\newcommand{\vg}{\mbox{$\myvec{g}$}}
\newcommand{\vh}{\mbox{$\myvec{h}$}}

\newcommand{\vm}{\mbox{$\myvec{m}$}}

\newcommand{\vs}{\mbox{$\myvec{s}$}}

\newcommand{\vw}{\mbox{$\myvec{w}$}}

\newcommand{\vx}{\mbox{$\myvec{x}$}}

\newcommand{\vy}{\mbox{$\myvec{y}$}}

\newcommand{\vz}{\mbox{$\myvec{z}$}}
\newcommand{\vA}{\mbox{$\myvec{A}$}}

\newcommand{\vC}{\mbox{$\myvec{C}$}}

\newcommand{\vE}{\mbox{$\myvec{E}$}}
\newcommand{\vF}{\mbox{$\myvec{F}$}}
\newcommand{\vG}{\mbox{$\myvec{G}$}}
\newcommand{\vH}{\mbox{$\myvec{H}$}}
\newcommand{\vI}{\mbox{$\myvec{I}$}}

\newcommand{\vK}{\mbox{$\myvec{K}$}}

\newcommand{\vM}{\mbox{$\myvec{M}$}}
\newcommand{\vN}{\mbox{$\myvec{N}$}}
\newcommand{\vO}{\mbox{$\myvec{O}$}}

\newcommand{\vS}{\mbox{$\myvec{S}$}}

\newcommand{\vW}{\mbox{$\myvec{W}$}}
\newcommand{\vX}{\mbox{$\myvec{X}$}}

\newcommand{\be}{\begin{equation}}
\newcommand{\ee}{\end{equation}}
\newcommand{\bea}{\begin{eqnarray}}
\newcommand{\eea}{\end{eqnarray}}
\newcommand{\beaa}{\begin{eqnarray*}}
\newcommand{\eeaa}{\end{eqnarray*}}

\begin{document}

\newcommand{\ourshorttitle}{
Can We Remove the Square-Root in Adaptive Gradient Methods?
}
\newcommand{\ourtitle}{
\vspace{-0.2cm}
\ourshorttitle \\ A Second-Order Perspective
\vspace{-0.2cm}
}
\icmltitlerunning{\ourshorttitle}

\twocolumn[

\icmltitle{\ourtitle}

\begin{icmlauthorlist}
\icmlauthor{Wu Lin}{vector}
\icmlauthor{Felix Dangel}{vector}
\icmlauthor{Runa Eschenhagen}{cambridge}
\icmlauthor{Juhan Bae}{ut}
\icmlauthor{Richard E. Turner}{cambridge}
\icmlauthor{Alireza Makhzani}{vector,ut}
\end{icmlauthorlist}

\icmlaffiliation{vector}{Vector Institute, Canada}
\icmlaffiliation{cambridge}{Cambridge University, United Kingdom}
\icmlaffiliation{ut}{University of Toronto, Canada}

\icmlcorrespondingauthor{Wu Lin}{yorker.lin@gmail.com \vspace{-0.2cm}}

\icmlkeywords{Natural Gradient Descent, Second Order Method, Optimization, Deep Learning}

\vskip 0.3in
]

\printAffiliationsAndNotice{}  %

\begin{abstract}
  Adaptive gradient optimizers like Adam(W) are the default training algorithms for many deep learning architectures, such as transformers.
Their diagonal preconditioner is based on the gradient outer product which is incorporated into the parameter update via a square root.
While these methods are often motivated as approximate second-order methods, the square root represents a fundamental difference.
In this work, we investigate how the behavior of adaptive methods changes when we remove the root, i.e.\,strengthen their second-order motivation.
Surprisingly, we find that such square-root-free adaptive methods \emph{close} the generalization gap to SGD on convolutional architectures, while \emph{maintaining} their root-based counterpart's performance on transformers.
The second-order perspective also has practical benefits for developing non-diagonal methods that can incorporate arbitrary curvature approximations through the concept of \emph{preconditioner invariance}.
In contrast to root-based methods like Shampoo, root-free counterparts work well and fast with half-precision since they do not require numerically unstable matrix root decompositions and inversions. 
Overall, our findings provide new insights into the development of adaptive methods and raise important questions regarding the overlooked role of adaptivity in their success.
\end{abstract}

\section{Introduction}
\begin{figure*}
  \centering
  \includegraphics[width=\linewidth]{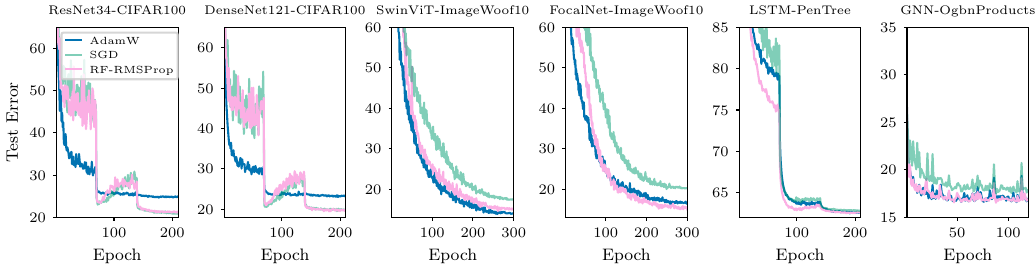}
  \vspace{-6ex}
  \caption{In modern (pre-)training setups (learning rate schedule, random search using 200 runs), square-root-free (RF) adaptive methods close the generalization gap between their root-based counterparts and SGD on CNNs (CIFAR-100), while maintaining their performance on vision transformers (ImageWoof10). They work well on other problems, like training a 3-layer LSTM, and a GNN with attention~\citep{zhang2022graph}.
  Experimental setup, performance measurements, and fine-tuning experiments on vision models are described in \Cref{sec:app_exp_nn}.
  \vspace{-0.4cm}
  }
  \label{fig:visual-abstract}
\end{figure*}

Adaptive gradient-based methods like Adam \citep{kingma2014adam} play a significant role in training modern deep learning models such as transformers.
A better understanding of these methods allows us to address their shortcomings and develop new adaptive methods to reduce training time and improve performance.
This is essential as deep learning models become increasingly large and complex to train.

Despite their success on architectures like transformers, adaptive methods tend to generalize worse than stochastic gradient descent (SGD) on convolutional architectures \citep{wilson2017marginal}.
Our understanding of this discrepancy is limited.
\citet{balles2018dissecting} dissect Adam into two concepts, sign descent and adaptive step sizes, and hypothesize that the connection to sign descent causes the gap to SGD on CNNs.
Similarly, \citet{kunstner2023noise,chen2023symbolic} argue that adaptive methods outperform SGD on transformers due to their connection to sign descent.  

It is challenging  to isolate the sign descent component of adaptive methods like Adam or RMSProp \citep{tieleman2012rmsprop}, as they typically introduce a square root to the preconditioner, which conflates the sign and adaptivity aspects and
hinders our comprehension of the role of the adaptivity.
The root is motivated by reports of improved performance \citep{tieleman2012rmsprop} and to stabilize convergence near the optimum \citep{kingma2014adam,kunstner2019limitations,martens2020new}.
However, it conflicts with the motivation of adaptive methods as approximate second-order methods based on the empirical Fisher, which is also commonly mentioned in works introducing adaptive methods~\citep[e.g.][]{kingma2014adam}.

Here, we investigate how the behavior of adaptive methods changes when we remove the root. Our idea is to strengthen the often-mentioned link to second-order methods that is weakened by the root.
Conceptually, this cleanly disentangles the aforementioned adaptivity aspect from the sign aspect.
Practically, it provides an opportunity to revisit the root's role in the context of modern training strategies, like non-constant learning rate schedules \citep{loshchilov2016sgdr} and hyperparameter tuning schemes \citep{choi2019empirical}, that differ from the original pipelines in which the square root was introduced.
Computationally, removing the root is beneficial for bridging the computation gap (\Cref{fig:experiment-matrix}) between diagonal and non-diagonal \emph{matrix} adaptive methods and lowering the per-iteration cost of matrix methods by removing matrix root decompositions that need to be carried out in high precision to avoid numerical instabilities \cite{gupta18shampoo,anil2020scalable}.

There are some challenges to establishing a rigorous second-order perspective on adaptive methods; one cannot just remove the root. We overcome those challenges and make the following contributions:
\begin{itemize}
\item We establish a rigorous second-order view of adaptive methods: we remove the root (Sec.~\ref{sec:first_order_view}), show how to interpret the  gradient outer product as a \emph{new} empirical Fisher variant (Sec.~\ref{sec:second_order}), adjust the preconditioner initialization (Sec.~\ref{sec:matrix}), and emphasize the importance of incorporating mini-batch curvature approximations (e.g., the outer products) from previous iterations (Secs.~\ref{sec:first_order_view},\ref{sec:second_order},\ref{sec:matrix}).
\item
Empirically, we show that---surprisingly---removing the root not only \emph{closes} the generalization gap between adaptive methods and SGD on convolutional neural networks, but \emph{maintains} the performance of square-root-based methods on vision transformers (Sec.~\ref{sec:first_order_view}).
\item
Conceptually, we introduce \emph{preconditioner invariance} (Sec.~\ref{sec:matrix}) through the second-order view to incorporate \emph{arbitrary} curvature approximations, remove inverses in root-free matrix adaptive methods, and 
bridge the computation gap between diagonal and matrix adaptive methods by developing new inverse- and root-free matrix methods that can train in low precision (Fig.~\ref{fig:experiment-matrix}). 
\item Consequently, we propose root-free RMSProp (Fig.~\ref{fig:rmsprop}) and root- and inverse-free Shampoo (Fig.~\ref{fig:matrix_methos_simple}) which are invariant to scaling the loss and affine reparametrization (Sec.~\ref{sec:second_order}), and work well on a variety of models (CNNs, LSTMs, GNNs, ViTs, and VMambas).
\end{itemize}

We release IF-Shampoo's PyTorch implementation at
\url{https://github.com/f-dangel/sirfshampoo}.
Our results can be generated using this repository: \url{https://github.com/yorkerlin/remove-the-square-root}.

\begin{figure*}[tb]
  \centering
  \includegraphics[width=\linewidth]{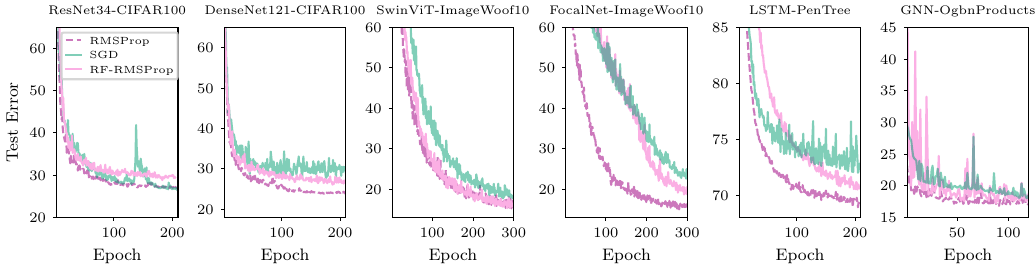}
  \vspace{-5ex}
  \caption{
  Comparison of root-based versus square-root-free (RF) methods in the original (outdated) training setup the root was introduced in (see \Cref{sec:app_exp_nn} for details).
  Adaptive methods with the root work better than their root-free counterparts when using (1) a constant learning rate schedule, (2) default zero initialization for a preconditioner, (3) default scaling for an averaged mini-batch loss. \vspace{-0.35cm}
  }
  \label{fig:experiment-constant-lr-schedule}
\end{figure*}

\section{First-order View of Adaptive Methods}
\label{sec:first_order_view}
For many deep learning tasks, training a neural network (NN) means solving an unconstrained optimization problem. For simplicity, consider a supervised learning task with a set of $N$ data points $\{y_i, \vx_i \}_{i=1}^{N}$, where  $y_i$ and $\vx_i$ represent a label and a feature vector. Given a NN $f(\cdot;\vmu)$ with learnable weights $\vmu$, the optimization problem is
 \begin{equation}
 \label{eq:optobj}
    \resizebox{0.6\hsize}{!}{%
    $
 \min_{\mu} \ell(\vmu) \coloneqq   \sum_{i=1}^{N} c(y_i, f(\vx_i; \vmu))\,,
    $%
}
\end{equation}
where $\hat{y}_i \coloneqq f(\vx_i;\vmu)$ is a predicted label of a feature vector $\vx_i$ and $c(y_i, \hat{y}_i)$ is a loss function that measures the discrepancy between a true ($y_i$) and predicted ($\hat{y}_i$) label.

To minimize \eqref{eq:optobj}, we can use adaptive gradient methods, which use the following preconditioned gradient update
 \begin{equation}
 \label{eq:adaupdate}
    \resizebox{0.4\hsize}{!}{%
    $
 \vmu \leftarrow \vmu - \stepsize_1 \vS^{-1} \nabla_\mu \ell(\vmu)\,,
    $%
}
\end{equation} 
where $\stepsize_1$ is an (initial) learning rate, $\nabla_\mu \ell(\vmu)$ is a gradient vector, and $\vS$ is a preconditioning matrix.
When $\vS$ is the Hessian and $\stepsize_1=1$, this becomes Newton's method.
In adaptive gradient methods, the preconditioning matrix $\vS$ is estimated by using only gradient information.

To estimate the preconditioner $\vS$, many adaptive methods employ an outer product $\vH  \coloneqq \vg  \vg^\top$ of a gradient vector $\vg$ that is usually the mini-batch gradient of \Cref{eq:optobj}:
\emph{Diagonal} adaptive methods such as RMSProp \citep{tieleman2012rmsprop}, diag-AdaGrad \citep{duchi2011adaptive}, and Adam \citep{kingma2014adam}, only use the diagonal entries
$\mathrm{diag}(\vH)=\vg \odot \vg$ where $\odot$ denotes element-wise multiplication.
\emph{Full matrix} adaptive methods like full-AdaGrad \citep{duchi2011adaptive} and its variants \citep{agarwal2019efficient} use the gradient outer product $ \vH$.
\emph{Structured matrix} methods like Shampoo \citep{gupta18shampoo} use a Kronecker-factored approximation of the gradient outer product $\vH$.

All of these methods apply a square root to the preconditioner in their update, and we will refer to them as square-root-based methods.
For example, RMSProp updates
its preconditioner $\vS=\mathrm{diag}( \sqrt{\hat{\vs}})$ and parameters according to 
\begin{equation}
    \label{eq:rmsprop}
    \resizebox{0.9\hsize}{!}{%
    $
 \hat{\vs}  \leftarrow (1-\stepsize_2) \hat{\vs} + \stepsize_2 \vh\,,\,
 \vmu  \leftarrow  \vmu - \stepsize_1  \vS^{-1}\vg= \vmu - \stepsize_1 \frac{\vg}{ \sqrt{{ \hat{\vs}}}}\,,
    $%
}
\end{equation}
where $\vh \coloneqq \mathrm{diag}(\vH)$,  $\stepsize_1$ is the learning rate, and $\stepsize_2$ is the weight to incorporate new outer products.
In full-matrix cases, adding the root requires matrix root decomposition. For example, full-AdaGrad computes the matrix root for its preconditioner $\smash{\vS= \hat{\vS}^{\nicefrac{1}{2}}}$ and updates parameters as
\begin{equation}
\label{eq:mat_adagrad}
    \resizebox{0.6\hsize}{!}{%
    $
\hat{\vS} \leftarrow  \hat{\vS} +  \stepsize_2 \vg \vg^\top\,,\,  \vmu  \leftarrow \vmu - \stepsize_1 \vS^{-1} \vg\,.
    $%
}
\end{equation}

\vspace{-0.4cm}
\subsection{Benefits and Results of Removing the Square Root}
\vspace{-0.2cm}
It is unclear how \emph{exactly} the square root emerged in adaptive methods.
Here, we hypothesize how it got introduced, critically assess why it may be desirable to remove,
and highlight connections to our experimental results.
We find that few works study square-root-free adaptive methods, and that benefits of removing the root are often overlooked.
Our work fills in this gap, which we think is crucial to better understand these methods and
our empirical results underline the great potential of square-root-free adaptive methods.

\vspace{-0.15cm}
\paragraph{Empirical performance \& training schemes}
The most important motivation in favor of the square root is the strong empirical performance of square-root-based methods that has been demonstrated in various works and rightfully established them as state-of-the-art for training transformers.
However, the context in which the root was introduced significantly differed from the training schemes that are used nowadays.
Original works such as \citet{tieleman2012rmsprop} show that including the square root improves the performance of adaptive methods when only tuning a learning rate that is fixed throughout training \citep{bottou2018optimization}.
Beyond this original setting, square-root-based methods have demonstrated great capabilities to train a wide range of NNs, e.g.\,when using a non-constant learning rate \citep{loshchilov2016sgdr} and
random search \citep{bergstra2012random,choi2019empirical} to tune all available hyper-parameters.
We wonder whether the square root, while necessary to achieve good performance in outdated training schemes, might \emph{not} be required in contemporary schemes.
To investigate this hypothesis, we conducted an experiment that compares square-root-based and square-root-free methods using the original training scheme in which the square root was introduced (\Cref{fig:experiment-constant-lr-schedule}).
Indeed, we find that the square root is beneficial in the original context.

\vspace{-0.3cm}
\paragraph{Interpretability \& generalization}
Square-root-based methods are state-of-the-art for training attention-based models \cite{zhang2020adaptive}, but exhibit a generalization gap to SGD on CNNs \cite{wilson2017marginal}.
The square root complicates the understanding of this phenomenon since adding the root conflates sign descent and adaptivity.
The sign-based component is often considered as a salient feature of adaptive methods while adaptivity is not.
For example, recent studies attribute the superior performance of adaptive methods over SGD on attention models to their sign-based update rules \citep{kunstner2023noise,chen2023symbolic}.
On the other hand, \citet{balles2018dissecting} hypothesize that sign descent may cause poor generalization of Adam on CNNs.
However, they neither consider the direct effect of the square root nor remove it from an existing method.
Therefore, the role of adaptivity is overlooked when understanding the gap between adaptive methods and SGD.
Removing the square root could clarify this issue, as a root-free method no longer performs sign descent while preserving adaptivity.
To investigate the role of adaptivity, we experiment with root-free RMSProp (\Cref{fig:rmsprop}) on attention and convolutional models (\Cref{fig:visual-abstract}).
For attention architectures, we find that removing the root does \emph{not} harm the performance of RMSProp.
Root-free methods with update clipping \citep{liu2023sophia} also perform well on large transformer-based language models.
These findings suggest that adaptivity is equally important as sign descent to explain the performance gap between adaptive methods and SGD on transformers.
For convolutional architectures, removing the root closes the generalization gap between RMSProp and SGD. This suggests that root-free adaptive methods can generalize well, and raises novel questions on the understanding of the role of adaptivity. %

\begin{figure*}[tb]
  \centering
  \includegraphics[width=\linewidth]{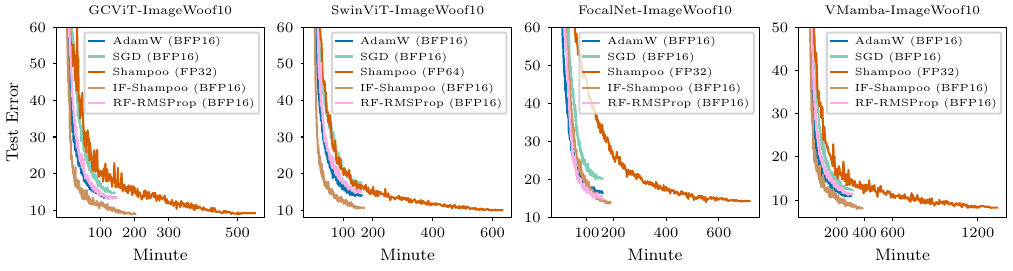}
  \vspace{-6ex}
  \caption{
  Comparison of matrix root-free versus root-based methods on GCViT~\citep{hatamizadeh2023global}, SwinViT~\citep{liu2021swin},  FocalNet~\citep{yang2022focal}, and VMamba~\citep{liu2024vmamba}. 
  Both matrix methods (Shampoo, IF-Shampoo) outperform diagonal methods on modern models and achieve a lower test error using modern training strategies (random search using 200 runs).
  In contrast to Shampoo, our inverse-free matrix  method, IF-Shampoo, runs in BFP-16 and trains twice as fast, while using less memory. 
  Using low-precision data types bridges the computation gap between diagonal and matrix methods. 
  All models are trained for 300 epochs. We update 
  matrix preconditioners at every 2 iterations and 
  can reduce the wall clock time by updating them less frequently.
  See App.~\ref{sec:app_exp_nn} for more details.
\vspace{-0.4cm}
  }
  \label{fig:experiment-matrix}
\end{figure*}

\vspace{-0.25cm}
\paragraph{Computational cost}
As noted by \citet{duchi2011adaptive}, the root poses numerical and computational challenges on \emph{matrix}  methods like Eq.~\eqref{eq:mat_adagrad} as it requires matrix roots which must be carried out in high precision to avoid numerical instabilities. 
This increases the run time and memory footprint and complicates the implementation \citep{anil2020scalable,shi2023distributed}.
Using low-precision data types \citep{micikevicius2017mixed} is a key technique to boost training speed and reduce memory.
The square root makes matrix methods undesirable for modern low-precision training schemes even when we can use iterative solvers \citep{anil2020scalable} to compute matrix roots.
By removing the matrix roots, and using advances on second-order methods \cite{lin2023simplifying}, we develop inverse-free matrix methods (Sec.~\ref{subsec:better_shampoo}) that are suitable for mixed-precision training.
We present empirical results for the low-precision setting in Fig.~\ref{fig:experiment-matrix}.
By removing the roots, we can consistently train in BFP-16, whereas root-based matrix methods like Shampoo need single---sometimes even double---precision.
On modern vision models, we find that root-free matrix methods outperform diagonal methods and perform similarly to Shampoo (c.f. Fig.~\ref{fig:experiment-matrix-more}.) while requiring less time and memory (c.f. Fig.~\ref{fig:experiment-matrix}) due to low-precision training.
This shows that removing the roots allows us to overcome the challenges of existing matrix methods, bridge the computation gap to diagonal methods, and expand their applicability to modern training pipelines for large models.

\vspace{-0.35cm}
\paragraph{Theoretical considerations}
\emph{Invariances.} Adding the root makes a descent step invariant to the scale of the loss -- the square root adjusts the scale of the ``squared'' gradient to be consistent with the gradient when the loss function is scaled.
This is useful as there is no need for users to pay attention to whether the loss function is averaged or summed over data points.
While adding the square root fixes the scaling issue, it breaks the affine reparameterization invariance \citep{nesterov1994interior}.
We can fix the scaling issue and preserve the affine invariance without the square root, as will be shown in \Cref{sec:second_order}; see
\Cref{sec:app-example-affine-invariance} for an example of the affine invariance of square-root-free methods.
Moreover, removing the root makes it easy to
introduce a new kind of invariance for preconditioning matrices (\Cref{sec:matrix}).

\vspace{-0.1cm}
\emph{Convergence analysis.} Theoretical works such as \citet{duchi2011adaptive,reddi2019convergence} and many others suggest that adding the square root is useful to prove convergence bounds for convex loss functions.
Convergence analysis for root-based methods is also extended to non-convex settings under certain assumptions such as gradient Lipschitz and Polyak-Łojasiewicz conditions.
However, compared to the regret bound of AdaGrad \citep{duchi2011adaptive} studied in convex settings,
\citet{Hazan2006LogarithmicRA} give a better regret bound in strongly-convex cases \emph{without} introducing the square root.
\citet{crammer2009adaptive} also study root-free methods in convex cases.
Recent works like \citet{Mukkamala2017VariantsOR,WangLCT020} prove similar bounds for root-free methods on convex problems. Thus, root-free methods are theoretically grounded---at least in convex settings.

\vspace{-0.1cm}
\emph{Behavior near optimum.} The square root is often introduced to avoid oscillation near an optimal solution where the preconditioner can be ill-conditioned.
For example, in one dimension, the descent direction $ s^{-1} g$ can be unbounded near the optimum when we use the outer product as the preconditioner $s=g^2$. This is 
because the gradient $g$ is close to $0$ near the optimum and the outer product as a `squared' gradient decreases much faster than the gradient $g$.
Thus, the update without the square root can lead to oscillation when using a constant learning rate.
However, even without the square root, a preconditioner can still be well-conditioned when incorporating outer products from previous iterations \citep{roux2007topmoumoute} and using Tikhonov damping \citep{becker1988improving,martens2012training}.
For example, the descent direction $ s^{-1} g$  can be bounded without the square root even when  near an optimal solution, where $g$ is a gradient and $s= (1-\stepsize_2) s + \stepsize_2 g^2$ is a preconditioner estimated by an exponentially moving average when $0<\stepsize_2<1$.

\begin{figure*}[!t]
\center
	   \fbox{
			\begin{minipage}{.3\textwidth}
		\textbf{RMSProp}

				\begin{algorithmic}[1]
            \STATE
            \footnotesize  Compute gradient $\vg\coloneqq\nabla \ell_{\text{scaled}}(\vmu)$  \\
             \scalebox{0.9}{   $
            \hat{\vs}  \leftarrow (1-\stepsize_2) \hat{\vs} + \stepsize_2 \vg^2
             $} \\
            \STATE
          \scalebox{0.9}{    $
          \vmu \leftarrow \vmu  - \stepsize_1 \vg /  \sqrt{\hat{\vs}}
 $}
				\end{algorithmic}
	\end{minipage}
	}
	\fbox{
			\begin{minipage}{.3\textwidth}
             \textbf{Square-root-free RMSProp (Ours) }

		\begin{algorithmic}[1]
               \STATE
            \footnotesize  Compute gradient $\vg\coloneqq\nabla  \ell_{\text{scaled}}(\vmu)$  \\
             \scalebox{0.9}{   $
            \vs  \leftarrow (1-\stepsize_2) \vs + \stepsize_2  {\color{red} B}\vg^2
             $} \\
            \STATE
          \scalebox{0.9}{    $
          \vmu \leftarrow \vmu  - \stepsize_1  \vg /   \vs
 $}
				\end{algorithmic}
	\end{minipage}
	}
 \vspace{-0.1cm}   \caption{
 Diagonal adaptive methods for a (scaled) loss function $\ell_{\text{scaled}}(\vmu)$ defined by averaging over $B$ data points in a mini-batch case.
 The scalar $B$ highlighted in red is essential because we \emph{average} the loss functions for mini-batch training.
 Hyperparameters will be hard to tune if the scalar is not included. This is because they will implicitly depend on the batch size $B$.
We initialize $s$ to $1$  in our root-free method while $\hat{s}$ is initialized to $0$ in the original RMSProp.
For simplicity, we do not include damping, weight decay, and momentum. A full-fledged version is in the Appendix, Fig.~\ref{fig:rmsprop-full}.
\vspace{-0.3cm}
}\label{fig:rmsprop}
\end{figure*}

\vspace{-0.2cm}
\section{A Second-order Perspective}
\label{sec:second_order}

\vspace{-0.2cm}
Here, we describe a second-order perspective on adaptive methods when the root is removed. This naturally resolves the scaling issue and is coherent with the original motivation to develop these methods.

Without the square root, an adaptive method takes a step
\begin{equation}
 \label{eq:root_free_update}
    \resizebox{0.7\hsize}{!}{%
    $
 \vS \leftarrow (1- \stepsize_2 \gamma) \vS + \stepsize_2 \vH\,,\qquad \vmu \leftarrow \vmu - \stepsize_1 \vS^{-1} \vg\,,
    $%
}
\end{equation}

\vspace{-0.3cm}
 where $\gamma \in \{0,1\}$ and the outer product $\vH=\vg\vg^\top$ is used as a curvature approximation.
Note that we incorporate outer products from previous iterations as $\stepsize_2 >0$.
For example, when $\gamma=0$ and $\vS$ is a full matrix, we obtain the update of full-AdaGrad without the square root.

The above update resembles a second-order method.
However, it is not invariant to the scale of the loss, unlike Newton's method: Scaling the loss by a constant $c$ scales the gradient and Hessian by $c$, but the gradient outer product by $c^2$, which is inconsistent with its role as a Hessian approximation and therefore conflicts with the second-order interpretation.
We will resolve this particular conflict and improve the justification of square-root-free adaptive methods through an approximate second-order view.

The key step is to view the outer product as a novel empirical Fisher that \emph{differs} from the standard empirical Fisher discussed in the DL literature \citep{kingma2014adam,kunstner2019limitations,martens2020new}.
Our empirical Fisher relies on the aggregated mini-batch gradient rather than per-sample gradients.
Since the Fisher is tied to a probability distribution that must be normalized, this provides a gauge to make updates invariant to the scale of the loss automatically.

\subsection{New Fisher Matrices as Hessian Approximations}
\vspace{-0.15cm}
Recall the objective function \scalebox{0.78}{ $\ell(\vmu)=\sum_{i=1}^{N}c(y_i, f(\vx_i; \vmu))$ } defined in \eqref{eq:optobj}.
Given a datum $\vx_i$, we can define a per-example probability distribution over label $y_i$ as \scalebox{0.78}{ $p(y_i\mid\vx_i; \vmu)=\exp(- c(y_i, f(\vx_i; \vmu)) )$}  by using the loss function $c(y_i, f(\vx_i; \vmu))$ (e.g., cross-entropy or square loss).
We denote an individual gradient for datum $\vx_i$ by \scalebox{0.78}{ $\vg_i\coloneqq\nabla_\mu c(y_i, f(\vx_i; \vmu))=-\nabla_\mu \log p(y_i|\vx_i;\vmu) $ }.

{\bf (1) Standard empirical Fisher matrices}
\begin{defn}
The empirical Fisher information (FIM) matrix \citep{pascanu2013revisiting,kingma2014adam,kunstner2019limitations,martens2020new}  $ \tilde{\vF}_{\text{standard}}(\vmu) $ is defined by replacing the expectation in the standard FIM $\vF_{\text{standard}}(\vmu)$ with observed labels.
\vspace{-0.3cm}
  \begin{equation*}
    \resizebox{0.95\hsize}{!}{%
    $
    \begin{aligned}
       {\vF}_{\text{standard}}(\vmu)  & \coloneqq \sum_{i=1}^{N}   \mathbb{E}_{ y_i \sim p(y_i\mid\mathbf{x}_i; \vmu) } \big[  \nabla_\mu \log p(y_i\mid\vx_i; \vmu) \nabla_\mu^\top \log p(y_i\mid\vx_i; \vmu) \big]  \\
    \tilde{\vF}_{\text{standard}}(\vmu)  & \coloneqq \sum_{i=1}^{N} \nabla_\mu \log p(y_i\mid\mathbf{x}_i; \vmu) \nabla_\mu^\top \log p(y_i\mid\vx_i; \vmu) =   \sum_{i=1}^{N} \vg_i \vg_i^\top
    \end{aligned}
    $%
}
\end{equation*}
\end{defn}

\vspace{-0.3cm}
Now, we introduce a new Fisher matrix as a FIM over a \emph{joint} distribution of labels.

{\bf (2) Our Fisher matrices for the original (unscaled) loss  }
\begin{defn}
\label{def:fim}
Our Fisher matrix is defined as
  \begin{equation}\label{eq:fim_gg}
    \resizebox{0.9\hsize}{!}{%
    $
    \begin{aligned}
    \vF_{\text{new}}(\vmu) &  \coloneqq  \mathbb{E}_{\mathbf{y} \sim p(\mathbf{y}\mid\mathbf{X}; \vmu) } \big[ \nabla_\mu \log p(\vy\mid\vX; \vmu) \nabla_\mu^\top \log p(\vy\mid\vX; \vmu) \big] \\
    &= -\mathbb{E}_{\mathbf{y} \sim p(\mathbf{y}\mid\mathbf{X}; \vmu) } \big[ \nabla_\mu^2 \log p(\vy\mid\vX; \vmu) \big],
    \end{aligned}
    $%
}
\end{equation} where the labels $\vy=(y_1,\dots,y_N)$ are considered jointly as a random vector,
 $\smash{p(\vy|\vX; \vmu)\coloneqq\prod_{i=1}^{N} p(y_i\mid\vx_i; \vmu)}$ is its joint distribution, and $\vX=(\vx_1,\dots,\vx_N)$ is a feature matrix.
Importantly, the joint distribution must be normalized.
\end{defn}
\begin{figure*}[tb]
  \centering
  \includegraphics[width=\linewidth]{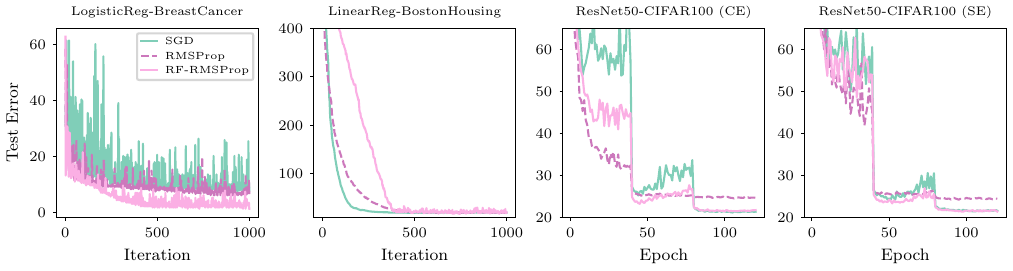}
  \vspace{-6ex}
  \caption{Experiments  demonstrating that square-root-free adaptive methods work well with both cross entropy (CE) and square error (SE)
  losses. Thus, our (diagonal) empirical Fisher estimation used in our update scheme does not suffer from the limitations of the standard empirical Fisher \citep{kunstner2019limitations}.
  We train all models using mini-batches and use random search (200 runs) to tune these methods.  
 In the first two plots on the left, we consider
 convex problems using
   a constant learning rate schedule (classical training scheme) considered by \citet{kunstner2019limitations}.
In the remaining plots, we consider
    non-convex NN problems with a step decay learning rate schedule (modern training scheme) considered by \citet{wilson2017marginal}.
  We consider  ResNet50 models and train them for 120 epochs with mini-batch size 128.
    Due to the large number of classes on the CIFAR100 dataset,
  we employ SE loss functions suggested by \citet{hui2020evaluation} when it comes to using SE loss functions for classification tasks.
\vspace{-0.2cm}
  } \label{fig:experiment-convex}
\end{figure*}

\begin{defn}
Our empirical Fisher matrix is defined by replacing the expectation in $\vF_{\text{new}}(\vmu)$ with observed labels.
  \begin{equation} \label{eq:fim_appro_hess}
    \resizebox{0.85\hsize}{!}{%
    $
    \begin{aligned}
    \tilde{\vF}_{\text{new}}(\vmu) &  \coloneqq   \nabla_\mu \log p(\vy\mid\vX; \vmu) \nabla_\mu^\top \log p(\vy\mid\vX; \vmu) = \vH   \\
    &\approx - \nabla_\mu^2 \log p(\vy\mid\vX; \vmu) = \nabla_\mu^2 \ell(\vmu)\,,
    \end{aligned}
    $%
}
\end{equation} with \scalebox{0.78}{$ \nabla_\mu \log p(\vy\mid\vX; \vmu)=  \sum_{i} \nabla_\mu \log p(y_i\mid\vx_i;\vmu) = -\sum_{i} \vg_i =-\vg $}.
\end{defn}

From \eqref{eq:fim_appro_hess}, we see that the outer product $\vH=\vg \vg^\top$ coincides with this empirical Fisher matrix $\tilde{\vF}_{\text{new}}(\vmu)$.

\begin{defn}
In a mini-batch case, we can define a Fisher matrix for a mini-batch with $B$ data points as
  \begin{equation} \label{eq:mini_fim}
    \resizebox{0.9\hsize}{!}{%
    $
    \begin{aligned}
    \vF_{\text{mini}}(\vmu)\coloneqq\mathbb{E}_{\mathbf{y}_{\text{mini}} \sim p } \big[ \nabla_\mu \log p(\vy_{\text{mini}}\mid\vX_{\text{mini}}; \vmu) \nabla_\mu^\top \log p(\vy_{\text{mini}}\mid\vX_{\text{mini}}; \vmu) \big]
\end{aligned}
$%
}
\end{equation} where   $\vy_{\text{mini}}\coloneqq(y_1,\dots,y_B)$ is a label vector,
  $\vX_{\text{mini}}\coloneqq(\vx_1,\dots,\vx_B)$ a feature matrix for the mini-batch, and $p(\vy_{\text{mini}}\mid\vX_{\text{mini}}; \vmu)\coloneqq\smash{\prod_{i=1}^{B}} p(y_i\mid\vx_i, \vmu)$ the joint distribution over labels for the mini-batch. This distribution can be obtained by marginalizing unseen labels in the original joint distribution $p(\vy\mid\vX; \vmu)$ defined for the full-batch.
This Fisher is an unbiased estimation of our full-batch Fisher as the claim---proof in \Cref{sec:app-proofunbiasedness}---is stated below.
\end{defn}
\begin{claim}
\label{claim:unbiasedness}
Our mini-batch Fisher
$\nicefrac{1}{B}\vF_{\text{mini}}(\vmu)$ is an unbiased estimation of the full-batch Fisher $\nicefrac{1}{N}\vF_{\text{new}}(\vmu)$.
\end{claim}
When we replace the expectation with observed labels, we obtain our empirical Fisher for the mini-batch. Note that this empirical Fisher is not an unbiased estimator of the full-batch empirical Fisher.
However, we often do not consider the unbiasedness when we incorporate empirical Fisher matrices from previous iterations.
Moreover, our interpretation provides the direct link to the Hessian as we can view the outer product as a Hessian approximation (see \eqref{eq:fim_appro_hess}) when the product corresponds to our Fisher. This interpretation allows us to preserve the scale-invariance to the loss and the affine reparametrization invariance. %

{\bf (3) Our Fisher matrices for a scaled loss  }

Now, we consider a scaled loss function. Then, the outer product $\vH$ no longer coincides with a Fisher matrix.
As an example, we consider averaging a loss over $N>1$ data points, as it is often done in mini-batch settings.
The loss is
\vspace{-0.2cm}
\begin{equation*}
    \resizebox{0.8\hsize}{!}{%
    $
{\ell}_{\text{scaled}}(\vmu) \coloneqq \frac{1}{N} \ell(\vmu) = \frac{1}{N} \sum_{i=1}^{N}c(y_i,f(\vx_i; \vmu)),
    $%
}
\end{equation*}
with gradient \scalebox{0.78}{$\vg_{\text{scaled}}\coloneqq\nicefrac{1}{N} \sum_{i=1}^{N} \nabla_\mu c (y_i, f(\vx_i;\vmu))=\nicefrac{1}{N} \sum_{i=1}^{N} \vg_i$} and outer product  and \scalebox{0.78}{ $\vH_{\text{scaled}}\coloneqq\vg_{\text{scaled}} \vg_{\text{scaled}}^\top=\nicefrac{1}{N^2} \vH$}.

In this case, the empirical Fisher matrix should be defined using the same joint distribution as
  \begin{equation} \label{eq:fim_appro_hess_scaled}
    \resizebox{0.91\hsize}{!}{%
    $
    \begin{aligned}
     \tilde{\vF}_{\text{scaled}}(\vmu) &\coloneqq   \frac{1}{N} \nabla_\mu \log p(\vy\mid\vX; \vmu) \nabla_\mu^\top \log p(\vy\mid\vX; \vmu) = N \vH_{\text{scaled}}  \\
    &= \frac{1}{N} \tilde{\vF}_{\text{new}}(\vmu)
    \approx  - \frac{1}{N}  \nabla_\mu^2 \log p(\vy\mid\vX; \vmu) = \nabla_\mu^2 \ell_{\text{scaled}}(\vmu)\,.
    \end{aligned}
    $%
}
\end{equation}
According to \eqref{eq:fim_appro_hess_scaled},
the outer product $\vH_{\text{scaled}}$  no longer coincides with this empirical Fisher matrix $\tilde{\vF}_{\text{scaled}}(\vmu)$.
When using $\tilde{\vF}_{\text{scaled}}(\vmu)$ as the Hessian approximation, the update of a square-root-free adaptive method is
\begin{equation*}
    \resizebox{\hsize}{!}{%
    $
 \vS \leftarrow (1- \stepsize_2 \gamma) \vS + \stepsize_2 \overbrace{  \tilde{\vF}_{\text{scaled}}(\vmu)} ^{ = {\color{red} N} \vH_{\text{scaled}} },\,\,\, \vmu \leftarrow \vmu - \stepsize_1 \vS^{-1} \vg_{\text{scaled}}
    $%
}
\end{equation*}
This update is scale-invariant since gradient \scalebox{0.78}{$ \vg_{\text{scaled}} = \nicefrac{1}{N} \vg $ } and empirical Fisher \scalebox{0.78}{ $ \tilde{\vF}_{\text{scaled}}(\vmu)  = N \vH_{\text{scaled}} =  \nicefrac{1}{N} \vH $} scale identically.

\paragraph{Affine reparametrization invariance}
Our fix does not break the  affine invariance of a square-root-free method.
A square-root-free method is affine invariant when using our scaled empirical Fisher matrix (see  Appx.~\ref{sec:app-proofaffine-invariance} for a proof).
\begin{claim}
\label{claim:affine_inv}
Our square-root-free update is affine invariant. %
\end{claim}

By using our empirical Fisher matrices,
our square-root-free methods like Newton's method not only is scale-invariant when we scale a loss function but also  preserves the affine reparametrization invariance. Thus, we consider our methods as approximate second-order (quasi-Newton) methods.

\subsection{Difference to the Standard Empirical Fisher}
Existing works \citep{kingma2014adam,kunstner2019limitations,martens2020new} do not distinguish the standard empirical Fisher denoted by \scalebox{0.78}{$\tilde{\vF}_{\text{standard}}(\vmu)$} from the outer product $ \vH $ corresponding to our empirical Fisher \scalebox{0.78}{$\tilde{\vF}_{\text{new}}(\vmu) $}.
They differ when $N>1$ since
  \begin{equation}   \label{eq:fim_appro_hess_incorrect}
    \resizebox{0.87\hsize}{!}{%
    $
    \begin{aligned}
    \tilde{\vF}_{\text{standard}}(\vmu) =    \sum_{i=1}^{N}\left( \vg_i \vg_i^\top \right)  \neq \left( \sum_{i=1}^{N} \vg_i  \right)  \left( \sum_{j=1}^{N} \vg_j  \right)^\top  = \vH =     \tilde{\vF}_{\text{new}}(\vmu)\,.
    \end{aligned}
    $%
}
\end{equation}
The standard empirical Fisher on the left is not rank-one, while our empirical Fisher on the right is.

This misconception can impede our understanding of these methods.
The outer product $\vH$ corresponding to our empirical Fisher $\tilde{\vF}_{\text{new}}(\vmu)$ is different from the standard empirical Fisher considered by \citet{kunstner2019limitations}.
Indeed, some limitations in \citet{kunstner2019limitations} are due to the ill-conditioning issue mentioned in Sec.~\ref{sec:first_order_view}.
Importantly, we estimate a preconditioner $\vS$ and overcome the ill-conditioning issue by incorporating outer products from previous iterations (e.g., by a moving average) -- even in the full-batch setting as shown in \eqref{eq:root_free_update}.
Aggregating these previous outer products makes our approximation distinct from the approximation considered by \citet{kunstner2019limitations} where the authors do not make use of previous products.
As shown in Fig.~\ref{fig:experiment-convex}, our estimation works well in both convex examples constructed by \citet{kunstner2019limitations} and non-convex problems.
Furthermore, our estimation has been theoretically justified \citep{Hazan2006LogarithmicRA,Mukkamala2017VariantsOR,WangLCT020} for root-free methods in (convex) settings similar to \citet{kunstner2019limitations}.
Thus, our empirical Fisher does not suffer from  limitations of the standard empirical Fisher considered in \citet{kunstner2019limitations} when using our  Fisher in the preconditioner update scheme.

\vspace{-0.25cm}
\subsection{Disentangling the Outer Product from the Fisher}
\begin{figure*}[!t] %
  \center
    \fbox{
    \begin{minipage}{.3\textwidth}
      \textbf{Shampoo  }

      \begin{algorithmic}[1]
        \STATE
        \footnotesize  Compute $\vG\coloneqq\mathrm{Mat}( \nabla  \ell_{\text{scaled}}(\vmu) )$  \\
        \scalebox{0.85}{   $
          \hat{\vS}_C  \leftarrow (1-\stepsize_2 \gamma) \hat{\vS}_C + \stepsize_2   \vG  \vG^\top
          $} \\
        \scalebox{0.85}{   $
          \hat{\vS}_K  \leftarrow (1-\stepsize_2 \gamma) \hat{\vS}_K + \stepsize_2\vG^\top  \vG
          $} \\

        \STATE
        \scalebox{0.82}{    $
          \mathrm{Mat}(\vmu) \leftarrow \mathrm{Mat}(\vmu) - \stepsize_1 \hat{\vS}_C^{-1/4} \vG \hat{\vS}_K^{-1/4}
          $}
      \end{algorithmic}
    \end{minipage}
  }
  \fbox{
    \begin{minipage}{.3\textwidth}
      \textbf{Square-root-free Shampoo  (Ours) }

      \begin{algorithmic}[1]
        \STATE
        \footnotesize  Compute $\vG\coloneqq\mathrm{Mat}( \nabla  \ell_{\text{scaled}}(\vmu) )$  \\
        \scalebox{0.82}{   $
          \vS_C  \leftarrow (1-\stepsize_2 \gamma ) \vS_C + \frac{\stepsize_2}{d} ( {\color{red} B } \vG \vS_K^{-1} \vG^\top )
          $} \\
        \scalebox{0.82}{   $
          \vS_K  \leftarrow (1-\stepsize_2  \gamma) \vS_K + \frac{\stepsize_2}{p} (  {\color{red} B } \vG^\top \vS_C^{-1} \vG  )
          $} \\
        \STATE
        \scalebox{0.82}{    $
          \mathrm{Mat}(\vmu) \leftarrow \mathrm{Mat}(\vmu) - \stepsize_1 \vS_C^{-1} \vG \vS_K^{-1}
          $}
      \end{algorithmic}
    \end{minipage}
  }
    \fbox{
      \begin{minipage}{.31\textwidth}
        \textbf{Inverse\&root-free Shampoo  (Ours)}

        \begin{algorithmic}[1]
          \STATE
          \footnotesize  Compute $\vG\coloneqq\mathrm{Mat}( \nabla  \ell_{\text{scaled}}(\vmu) )$  \\
          \scalebox{0.76}{   $
            {\vC}  \leftarrow {\vC} \exp(-\frac{\stepsize_2}{2 d} ( {\color{red} B }{\vC}^\top \vG {\vK} {\vK}^\top \vG^\top {\vC}  - \gamma d\vI ) )
            $} \\
          \scalebox{0.76}{   $
            {\vK}  \leftarrow {\vK} \exp(-\frac{\stepsize_2}{2 p}( {\color{red} B } \vK^\top \vG^\top {\vC} {\vC}^\top \vG {\vK}  - \gamma  p \vI ) )
            $} \\

          \STATE
          \scalebox{0.82}{    $
           \mathrm{Mat}(\vmu) \leftarrow \mathrm{Mat}(\vmu)   - \stepsize_1 {\vC}{\vC}^\top \vG {\vK}{\vK}^\top
            $}

        \end{algorithmic}
      \end{minipage}
    }
  \vspace{-0.1cm}   \caption{
Structured matrix adaptive methods for a (scaled) loss function $\ell_{\text{scaled}}(\vmu)$ defined by averaging over $B$ data points in a mini-batch case.
For simplicity, we assume $\mathrm{Mat}(\vmu) \in \mathbb{R}^{p \times d} $ is a weight matrix in a layer.
     $\hat{\vS}_C, \vS_C, \vC \in \real^{p \times p}$  and $\hat{\vS}_K, \vS_K, \vK \in \real^{d \times d}$ are non-singular matrices.
     In the inverse-free method, we directly update $\vC$ and $\vK$, and approximate the matrix exponential $\exp(\vN)$ by its first-order truncation $\exp(\vN)\approx \vI + \vN$, where $\vC$ and $\vS_C$ are related as $\vS_C^{-1} = \vC\vC^\top$ (c.f. Claim~\ref{claim:shampoo_invariance_free}).
     $\vS_K$ and $\vK$ are  similarly related.
    We initialize each of $\vC$, $\vS_C$, $\vS_K$, and $\vK$ to an identity matrix in our methods while each of $\hat{\vS}_C$ and $\hat{\vS}_K$ is initialized to zero \citep{gupta18shampoo} .
For simplicity, we do not include damping, weight decay, and momentum. See Fig.~\ref{fig:matrix_methos_full}    in the Appendix for a full-fledged version.
\vspace{-0.3cm}
  }\label{fig:matrix_methos_simple}
\end{figure*}

At first glance,  the  outer product \scalebox{0.75}{$\vH_{\text{scaled}}$} coincides with another empirical Fisher  \scalebox{0.78}{ $\tilde{{\vF}}_{\text{incorrect}}(\vmu)= \nabla_\mu \log p_{\text{scaled}}(\vy\mid\vX; \vmu) \nabla_\mu^\top \log p_{\text{scaled}}(\vy\mid\vX; \vmu) $}  in the scaled case. This  Fisher is defined over a \emph{different} joint distribution \scalebox{0.78}{$p_{\text{scaled}}(\vy\mid\vX; \vmu)= \smash{\prod_{i=1}^{N}} p_{\text{scaled}}(y_i\mid\vx_i; \vmu)$} and depends on a \emph{scaled} per-sample probability distribution over label $y_i$ as \scalebox{0.78}{ $p_{\text{scaled}}(y_i\mid\vx_i; \vmu)=\exp(- { \color{red} \nicefrac{1}{N} } c(y_i, f(\vx_i; \vmu)) )$}.
However, this scaled distribution is not normalized.
Therefore, we cannot define a valid FIM that corresponds to the  outer product.
In other words, the outer product is not an empirical Fisher matrix.
The normalization condition of a per-sample distribution is often overlooked in the literature when considering a gradient outer product as an empirical Fisher.

\vspace{-0.3cm}
\section{Matrix Adaptive Methods}
\label{sec:matrix}
We develop a class of matrix adaptive methods without using matrix decompositions for training with mini-batches.
This introduces another challenge: a dense matrix-valued preconditioner may be too costly to store.
We address this by enforcing the preconditioner to be \emph{structured}; specifically, Kronecker-factored.
We use Eq.~\eqref{eq:root_free_update} to incorporate curvature information from previous iterations, especially for training with \emph{mini-batches}. This is needed as we want to approximate full-batch curvature information.
This curvature aggregation is also useful to
handle the ill-conditioning issue mentioned in Sec.~\ref{sec:first_order_view} when mini-batch curvature denoted by $\mathcal{H}$ is a gradient outer product $\vH$.
Unfortunately, Eq.~\eqref{eq:root_free_update} suggests that the optimizer's preconditioner $\vS$ must have the same structure as the curvature approximation $\mathcal{H}$, so that the structure is maintained under the update.
Thus, an additional projection step is needed when $\mathcal{H}$ has an incompatible structure. For instance,  the outer product structure in  $\mathcal{H}$ when $\mathcal{H}=\vH$  is incompatible with the Kronecker-factored structure in $\vS$.
It is common practice to introduce a \emph{sequential} inner loop to solve an optimization sub-problem at the projection step.  However, this increases the iteration cost. Prime examples are the Frank-Wolfe method and Newton-CG \citep{nocedal1999numerical}.
One idea to avoid using such an inner loop is to use a customized approximation such as Shampoo \citep{gupta18shampoo} on a case-by-case basis.
Instead, we consider a \emph{generic} approach to aggregate \emph{arbitrary} curvature information  \emph{without} an inner loop.

\vspace{-0.2cm}
\subsection{
Square-root-free Methods through Gaussian
Variational Approximations
}
\label{subsec:blr}
\vspace{-0.2cm}
We start with a dense preconditioner $\vS$ to introduce the concept of preconditioner invariance which enables us to efficiently `project' an arbitrary curvature approximation onto a flexible class of pre-conditioner parameterizations via the Bayesian learning rule \citep[BLR,][]{khan2017conjugate, khan18a, zhang2018noisy,lin2020handling,lin2021tractable,tan2021analytic,khan2023bayesian} and the chain rule.

As discussed in Sec.~\ref{sec:second_order}, if the gradient outer product coincides with our empirical Fisher, we can view it as a Hessian approximation. This view allows us to extend the BLR originally developed for Newton's method.
According to the BLR, we consider the preconditioner $\vS$ in \Cref{eq:root_free_update} as an inverse covariance of a Gaussian,   and the curvature approximation $\mathcal{H}$ as a partial derivative.
By viewing the preconditioner as the inverse covariance, we gain new reparametrization invariance for the preconditioner $\vS$ while preserving the affine invariance in $\vmu$ (c.f. Claim \ref{claim:affine_inv}). %
This invariance allows us to reparameterize the preconditioner and obtain an
equivalent update up to first-order accuracy.
One immediate application is to make
matrix adaptive methods such as square-root-free full-AdaGrad \emph{inverse-free} (see Eq.~\eqref{eq:inv_update}) by reparameterizing the preconditioner,
which is helpful for low-precision training. %
Our inverse-free scheme is generic for any curvature approximation (i.e., $\mathcal{H}\neq \vg\vg^T$).
Thanks to the invariance, we can later impose sparse structures on the preconditioner~\cite{lin2023structured}. This is flexible since preconditioner and curvature approximation can have independent structures.

Now, we describe the BLR. We consider a Bayesian problem formulation and solve a variational inference (VI) problem with a Gaussian variational approximation. In this setting, we consider NN weights as random variables and use a new symbol $\vw$ to denote these weights as they are no longer learnable. We refer to $\vmu$ and $\vS$ as the mean and the inverse covariance of the Gaussian $q(\vw\mid\vmu,\vS)$ over weights $\vw$.
This VI problem \citep{barber1997ensemble} (with $\gamma=1$) is
\begin{align}
 \min_{\mu, S \succ 0} \mathcal{L}(\vmu,\vS)\coloneqq \mathbb{E}_{w \sim q} \big[ \ell(\vw)  \big] - \gamma \mathcal{Q}_q\,, \label{eq:viopt}
\end{align}

\vspace{-0.3cm}
where $\ell(\vw)$ is the same loss defined in \eqref{eq:optobj}, $\vS \succ 0$ must be  positive-definite (PD), $\gamma $ is the same hyperparameter as in \eqref{eq:root_free_update},  $q(\vw\mid\vmu,\vS)$ is a Gaussian approximation with mean $\vmu$ and covariance $\vS^{-1}$, and $\mathcal{Q}_q \coloneqq \mathbb{E}_{w \sim q}[- \log q(\vw\mid\vmu,\vS)] = -\nicefrac{1}{2} \log \mathrm{det} (\vS)$ is the Gaussian's entropy.
When $\gamma=0$, problem in Eq.~\eqref{eq:viopt} becomes a variational optimization (VO) problem \citep{staines2012variational,khan2017variational} or a Gaussian adaptation problem\footnote{See \citet{lin2021tractable} for connections to gradient-free evolution methods \citep{wierstra2008natural,glasmachers2010exponential}.
} \citep{taxen1992gaussian,back1993overview}. Thus, the VI problem can be viewed as an entropy-regularized VO problem.

The FIM of the Gaussian $q(\vw\mid\vtheta)$ under parameter $\vtheta$
\vspace{-0.2cm}
\begin{equation}\label{eq:fim_gauss}
    \resizebox{0.6\hsize}{!}{%
    $
\begin{aligned}
 \vF_{\text{gauss}}(\vtheta)\coloneqq -\mathbb{E}_{w \sim q}[\nabla_\theta^2 \log q(\vw\mid\vtheta)]\,,
 \end{aligned}$%
}
\end{equation}

\vspace{-0.3cm}
 has a closed-form expression and is PD whenever $\vtheta$ is a valid parameterization. E.g., $\vtheta\coloneqq(\vmu,\vS)$ is a valid parameterization iff
the inverse covariance $\vS$ is PD.
This FIM should not be confused with other Fisher matrices in Sec.~\ref{sec:second_order}.

We exploit a Riemannian geometric structure of the probabilistic  (Gaussian) surrogate $q(\vw\mid\vtheta)$ and, thus, use natural gradient descent (NGD) to solve the problem in \Cref{eq:viopt}. An NGD step in parameterization $\vtheta$ is
\begin{equation} \label{eq:ngd_gauss}
    \resizebox{0.55\hsize}{!}{%
    $
\begin{aligned}
\vtheta \leftarrow \vtheta - \stepsize_1 (\vF_{\text{gauss}}(\vtheta))^{-1} \nabla_\theta \mathcal{L}\,.
\end{aligned}$%
}
\end{equation}

\vspace{-0.2cm}
\citet{khan18a} show that this update is equivalent to
\begin{equation}\label{eq:blr_gauss}
    \resizebox{0.85\hsize}{!}{%
    $
    \begin{aligned}
\vS & \leftarrow \vS + 2 \stepsize_1 \partial_{S^{-1}} \mathcal{L} = (1-\stepsize_1 \gamma) \vS + \stepsize_1 {\color{red} \mathbb{E}_{w \sim q}[\nabla_w^2 \ell(\vw)]}  \\
\vmu & \leftarrow  \vmu - \stepsize_1  \vS^{-1} \partial_\mu \mathcal{L} = \vmu -\stepsize_1 \vS^{-1}{ \color{red}  \mathbb{E}_{w \sim q}[\nabla_w \ell(\vw)] }\,,
    \end{aligned}$%
}
\end{equation}
where we use the following identities\footnote{
See \citet{lin2021tractable} for sampling-based estimations to remove the need of computing $\nabla_w \ell(\vw)$ and $\nabla_w^2 \ell(\vw)$. They are often used in gradient-free optimization and VI.
} for the Gaussian $q(\vw\mid\vtheta)$ \citep{opper2009variational,wu-report},
\begin{equation}  \label{eq:grad_gauss}
    \resizebox{0.88\hsize}{!}{%
 $\partial_\mu \mathcal{L} = \mathbb{E}_{w \sim q}[\nabla_w \ell(\vw)],\,\,2\partial_{S^{-1}} \mathcal{L} =\mathbb{E}_{w \sim q}[\nabla_w^2 \ell(\vw)]- \gamma \vS\,.
 $%
}
\end{equation}

To recover the root-free update in \Cref{eq:root_free_update}, we approximate the update in \Cref{eq:blr_gauss} with
\begin{equation}
\label{eq:adaptive_root_free}
    \resizebox{0.88\hsize}{!}{%
 $
  \vS \leftarrow (1-\stepsize_2 \gamma) \vS + \stepsize_2 \overbrace{  \mathcal{H} }^{ \approx {\color{red}\nabla_\mu^2 \ell(\vmu) }},\,\, \vmu \leftarrow \vmu - \stepsize_1 \vS^{-1} \overbrace{ \vg}^{={\color{red} \nabla_\mu \ell(\vmu)}},
  $%
}
\end{equation} by [1] using a delta approximation at $\vmu$ to approximate the expectations highlighted in red,  [2] replacing the Hessian with the outer product $\vH$ as it is a Hessian approximation (e.g., $\mathcal{H}=\vH$) in the unscaled case shown in Eq.~\eqref{eq:fim_appro_hess}, and [3] introducing an extra learning rate $\stepsize_2$ to compensate for the error of the curvature approximation.
By using the delta approximation, we can use the NGD update derived for the Bayesian problem in Eq.~\eqref{eq:viopt} to solve the non-Bayesian problem in Eq.~\eqref{eq:optobj}.
This formulation\footnote{See \citet{lin2023simplifying} for a (non-Bayesian) manifold optimization formulation, where they view a Gaussian approximation with a non-constant mean as a PD submanifold. Newton's method can be viewed as  Riemannian gradient descent on an augmented PD submanifold by embedding the Gaussian into the submanifold.
} unveils the Gaussian approximation hidden in square-root-free updates as the NGD update recovers the square-root-free method.
If the loss is scaled, we can replace the outer product with our empirical Fisher as a proper curvature approximation such as $\mathcal{H}= \tilde{\vF}_{\text{scaled}}$ (c.f. Sec.~\ref{sec:second_order}).
Moreover, we recover the update rule (Eq.~\eqref{eq:adaptive_root_free}) to incorporate
curvature information from previous iterations  (e.g., by an exponentially moving average when $\gamma=1$) for mini-batch training.

\vspace{-0.2cm}
\paragraph{Examples}
Many square-root-free adaptive gradient methods can be derived from this update rule when the curvature approximation is the outer product  (i.e., $\mathcal{H}=\vH$).
For example, \Cref{eq:adaptive_root_free} becomes the update of square-root-free full-AdaGrad  when $\gamma=0$:

\vspace{-0.5cm}
\begin{equation}
\label{eq:rf_adamat}
    \resizebox{0.65\hsize}{!}{%
 $
  \vS \leftarrow  \vS +  \stepsize_2 \vH ,\,\, \vmu \leftarrow \vmu - \stepsize_1 \vS^{-1} \vg\,.$%
}
\end{equation}
When $\vS=\mathrm{diag}(\vs)$ is diagonal, the update of \eqref{eq:adaptive_root_free} becomes

\vspace{-0.5cm}
\begin{equation*}
    \resizebox{0.88\hsize}{!}{%
 $
  \vs \leftarrow (1-\stepsize_2 \gamma) \vs + \stepsize_2 \mathrm{diag}(\vH) ,\,\, \vmu \leftarrow \vmu - \stepsize_1 \vS^{-1}  \vg\,.$%
}
\end{equation*}
We obtain the root-free RMSProp update when \scalebox{0.85}{$\gamma=1$}; likewise, the root-free diag-AdaGrad update when
\scalebox{0.85}{$\gamma=0$}.

\vspace{-0.2cm}
\paragraph{Non-zero initialization}
The preconditioner $\vS$ is often initialized to zero in adaptive methods.
An immediate consequence of the BLR is that $\vS$ should be initialized to a \emph{non-zero} value because we view it as an inverse covariance.
As shown in Fig.~\ref{fig:experiment-constant-lr-schedule}, this is important for the performance of root-free methods.
Moreover, the updated $\vS$ in Eq.~\eqref{eq:adaptive_root_free} is guaranteed to be PD when  $\vS$ is initialized to a PD matrix since the product $\mathcal{H}=\vH=\vg\vg^\top$ is positive semi-definite. When $\mathcal{H}$ is an arbitrary approximation, we can add a correction term obtained using higher-order Riemannian information of the surrogate $q(\vw|\vtheta)$ \citep{lin2020handling} to handle the PD constraint of $\vS$.

\vspace{-0.15cm}
\paragraph{Inverse-free (IF) update via preconditioner invariance}
The BLR preserves the scale-invariance to the loss and affine reparametrization invariance in $\vmu$.
The scale invariance comes from a proper Hessian approximation $\mathcal{H}$.
The affine invariance in $\vmu$
is  inherited from  \emph{invariance of natural-gradients} in the parameter space (e.g., $\vtheta=(\vmu,\vS)$) of the Gaussian.
Importantly, the BLR also introduces a new invariance to reparametrize preconditioner $\vS$  thanks to the same invariance of natural gradients.

By exploiting this preconditioner invariance,
we can easily make the root-free (RF) full-AdaGrad update inverse-free by performing NGD in another parameter space (e.g., $\veta\coloneqq(\vmu,\vS^{-1})$). As shown in App.~\ref{sec:app_inverse_free_adagrad}, the update of  $\vS^{-1}$ is

\vspace{-0.55cm}
\begin{equation}  \label{eq:inv_update}
    \resizebox{0.8\hsize}{!}{%
 $
  \vS^{-1} \leftarrow  \vS^{-1} -  \stepsize_2 \vS^{-1}\vH \vS^{-1} + {\color{red} \frac{\stepsize_2^2}{2} \vS^{-1} \vH \vS^{-1} \vH \vS^{-1} }\,,
  $%
}
\end{equation}

\vspace{-0.45cm}
where we introduce a correction term highlighted in red to satisfy the PD constraint of $\vS^{-1}$ \cite{lin2020handling}.
Note that we can use the Woodbury matrix identity to compute the exact inverse in RF-AdaGrad since $\mathcal{H}=\vH=\vg\vg^T$ has a low-rank structure.
However, our inverse-free update scheme (e.g., Eq.~\eqref{eq:inv_update}) can be applied to other curvature approximations even when the identity is not applicable (e.g, $\mathcal{H}\neq \vH$).
See Appx.~\ref{sec:app_matrix_invar_full} for a discussion about the exact inverse when $\mathcal{H}=\vH$.
As described in Claim \ref{claim:inv_free_full_adagrad} (see
Appx.~\ref{sec:app_matrix_invar_full} for a proof),
this update is equivalent to RF-AdaGrad.
This update is IF since it directly updates $\vS^{-1}$.
Moreover, it is scale and affine invariant  (c.f. Appx.~\ref{app:invariance_full_root_free}) like RF-AdaGrad.

\vspace{0.1cm}
\begin{claim}
\label{claim:inv_free_full_adagrad}
The update of Eq.~\eqref{eq:inv_update} is equivalent to Eq.~\eqref{eq:rf_adamat}
up to  first-order accuracy.
\end{claim}
This illustrates preconditioner invariance.
Thus, we use this invariance to incorporate mini-batch curvature in a reparametrized space (e.g., Eq.~\eqref{eq:inv_update} for $\vS^{-1}$ when $\mathcal{H}=\vH$).
See \citet{lin2023simplifying} for another inverse-free update rule by directly updating a Cholesky-like factor, $\vA$, of $\vS^{-1}$ (i.e., $\vS^{-1}=\vA\vA^T$).

\subsection{Decoupling Preconditioner \& Curvature }
\vspace{-0.2cm}
We further exploit the preconditioner invariance to allow
 preconditioner $\vS$ and curvature $\mathcal{H}$ to have their own structures.
From Eq.~\eqref{eq:blr_gauss}, we see that preconditioner $\vS$ is the inverse covariance  while the curvature approximation $\mathcal{H}  $ appears as a term of partial derivative $2\partial_{S^{-1}} \mathcal{L}\approx \mathcal{H} -\gamma \vS $ due to Eq.~\eqref{eq:grad_gauss}.
If a sparse pattern in $\vS$ can be obtained via reparameterization, we can perform NGD in this reparameterized space.
This invariance then allows us to incorporate the curvature approximation $\mathcal{H}$ and perform an exponential moving average on sparse $\vS$
even when the structure of $\mathcal{H}$ is incompatible with $\vS$. This is because we can use the chain rule to `project' $\mathcal{H}$ (as a partial derivative) onto the (sparse) reparameterized space of $\vS$.
Importantly, this neither introduces an inner loop nor significantly increases computational overhead.
We will see an example next.
For more examples, see \citet{lin2021snd}.

\subsection{Kronecker-factored Adaptive Methods}
\label{subsec:better_shampoo}
\vspace{-0.15cm}

Shampoo is a square-root-based Kronecker-factored method, where the inner (matrix) square roots are introduced due to a structural approximation and the outer (matrix) square root is inherited from full-AdaGrad.
Shampoo's preconditioner $\vS$ is coupled to its curvature approximation $\mathcal{H}_{\text{shampoo}} \equiv \vH= \vg \vg^\top$.
Thus, it approximates the outer product \scalebox{0.75}{$\vH \approx {\hat{\vS}}_C^{\nicefrac{1}{2}} \otimes {\hat{\vS}}_K^{\nicefrac{1}{2}}$} and uses \scalebox{0.75}{$\vS=\left(\hat{\vS}_C^{\nicefrac{1}{2}} \otimes \hat{\vS}_K^{\nicefrac{1}{2}} \right)^{\nicefrac{1}{2}} $} as preconditioner to ensure they have the same structure. %
Because an exponential moving average on $\vS$ is infeasible due to the structural incompatibility between $\vH$ and $\vS$, Shampoo uses exponential moving averages over $\hat{\vS}_C$ and $\hat{\vS}_K$ as heuristics.
Consequently, its update
\begin{equation*}
    \resizebox{0.97\hsize}{!}{%
    $
    \begin{aligned}
 \hat{\vS}_C & \leftarrow (1-\stepsize_2 \gamma ) \hat{\vS}_C  +  \stepsize_2 \vG \vG^\top,\,\,\,  \hat{\vS}_K \leftarrow  (1-  \stepsize_2 \gamma ) \hat{\vS}_K + \stepsize_2   \vG^\top \vG \\
 \vmu & \leftarrow \vmu - \stepsize_1 \vS^{-1} \vg \iff \vM \leftarrow \vM  - \stepsize_1 {\hat{\vS}}_C^{-1/4} \vG {\hat{\vS}}_K^{-1/4},
    \end{aligned}
    $%
}
\end{equation*}

\vspace{-0.35cm}
does not generalize to curvature approximations different from $\vg \vg^\top$,
where $\vG\coloneqq\mathrm{Mat}(\vg)$ and $\vM\coloneqq\mathrm{Mat}(\vmu)$.

In contrast, we consider a structural reparameterization of the inverse covariance $\vS=\vS_C \otimes \vS_K$ as the preconditioner and perform NGD to update $\vS_C$ and $\vS_K$.
We treat a curvature approximation $\mathcal{H}$, such as the gradient outer product $\vH$, as a partial derivative and use it to update $\vS_C$ and $\vS_K$ by the chain rule.
We do not require the approximation $\mathcal{H}$ to have the same structure as the preconditioner $\vS$. Thus, we decouple the preconditioner  from the curvature approximation. Notably, our approach is root-free and does not require $\mathcal{H}$ to be an outer product---a key assumption of Shampoo.
For example, we can use other inexpensive approximations $\mathcal{H}$ such as KFAC \cite{martens2015optimizing}.

Concretely, consider a parameterization $\vtau\coloneqq(\vmu,\vS_C,\vS_K)$ of the Gaussian, where $\vS_C \in \real^{p \times p}$ and $\vS_K \in \real^{d \times d}$ are PD. This Gaussian is known as a matrix Gaussian\footnote{We generalize our update to $N$-dimensional tensors using a tensor Gaussian \citep{ohlson2013multilinear}. See our PyTorch implementation and \Cref{sec:app_inverse_and_rootfree_shampoo_Nd} for the details.
}.
The exact FIM  \scalebox{0.72}{$\vF_{\text{gauss}}(\vtau)=\begin{bmatrix}
\vF_{\mu\mu} & \mathbf{0} & \mathbf{0}\\
\mathbf{0} & \vF_{CC} & \vF_{CK} \\
\mathbf{0} & \vF_{KC} & \vF_{KK}
\end{bmatrix}$} under this parameterization is singular since the Kronecker product is not unique.
We consider a block-diagonal approximated FIM of the Gaussian denoted by
 \scalebox{0.72}{
$
    \begin{aligned}
\tilde{\vF}_{\text{gauss}}(\vtau)=\begin{bmatrix}
\vF_{\mu\mu} & \mathbf{0} & \mathbf{0}\\
\mathbf{0} & \vF_{CC} & \mathbf{0} \\
\mathbf{0} & \mathbf{0} & \vF_{KK}
\end{bmatrix}
    \end{aligned}
    $} without requiring $\mathcal{H}$ to be an outer product.
This approximate FIM is non-singular and well-defined by ignoring cross-terms $\vF_{CK}$ and $\vF_{KC}$ in the original FIM.
We propose to perform approximate NGD with \scalebox{0.8}{ $\tilde{\vF}_{\text{gauss}}(\vtau)$}:

\vspace{-0.6cm}
\begin{equation*}
    \resizebox{0.9\hsize}{!}{%
    $
    \begin{aligned}
    \begin{bmatrix}
 \vmu \\ \vS_C \\ \vS_K
\end{bmatrix}
& \leftarrow
\begin{bmatrix}
     \vmu \\ \vS_C \\ \vS_K
\end{bmatrix} - \stepsize_1
\begin{bmatrix}
\vS
& \mathbf{0} & \mathbf{0} \\
\mathbf{0} & -\frac{d}{2} \frac{\partial \vS_C^{-1} }{\partial \vS_C } & \mathbf{0} \\
\mathbf{0} & \mathbf{0} & -\frac{p}{2} \frac{\partial \vS_K^{-1} }{\partial \vS_K }
\end{bmatrix}^{-1}\begin{bmatrix}
 \partial_\mu \mathcal{L}  \\  \partial_{S_C} \mathcal{L}
 \\  \partial_{S_K} \mathcal{L}
\end{bmatrix}  \\
&=
\begin{bmatrix}
\vmu -\stepsize_1 \vS^{-1} \partial_\mu \mathcal{L} \\
\vS_C +  \frac{2\stepsize_1}{d} \textcolor{blue}{\partial_{S_C^{-1}} \mathcal{L}} \\
\vS_K + \frac{2\stepsize_1}{p} \textcolor{blue}{\partial_{S_K^{-1}} \mathcal{L}}
   \end{bmatrix},
    \end{aligned}
    $%
}
\end{equation*}

\vspace{-0.3cm}
where
 $\vF_{\mu\mu} =\vS =\vS_C \otimes \vS_K$ ,
 $\vF_{CC} = -\frac{d}{2} \nicefrac{\partial \vS_C^{-1} }{\partial \vS_C }$ ,
$\vF_{KK} =  -\frac{p}{2} \nicefrac{\partial \vS_K^{-1} }{\partial \vS_K }$ are computed according to  Eq.\eqref{eq:fim_gauss} or the block-wise Bregman duality since the matrix Gaussian is a member of a multi-linear exponential family \citep{lin2019fast}.
Then, we can use the chain rule to `project' the curvature $\mathcal{H}$ as a term of derivatives highlighted in blue

\vspace{-0.45cm}
\begin{equation*}
    \resizebox{0.8\hsize}{!}{%
    $
    \begin{aligned}
    2 [ \textcolor{blue}{\partial_{S_C^{-1}} \mathcal{L}}]_{ij} & = 2 \mathrm{Tr} \big( \frac{\partial \vS^{-1}}{ \partial [ \vS_C^{-1}]_{ij} } \partial_{S^{-1}}\mathcal{L}\big) \approx \mathrm{Tr}\big( \frac{\partial \vS^{-1}}{ \partial [\vS_C^{-1} ]_{ij} } [\textcolor{blue}{\mathcal{H}} - \gamma \vS_C \otimes \vS_K] \big) \\
  2[\textcolor{blue}{\partial_{S_K^{-1}} \mathcal{L}}]_{ij} &= 2 \mathrm{Tr}\big( \frac{\partial \vS^{-1}}{ \partial [\vS_K^{-1}]_{ij} } \partial_{S^{-1}}\mathcal{L} \big) \approx  \mathrm{Tr}\big( \frac{\partial \vS^{-1}}{ \partial [\vS_K^{-1} ]_{ij} } [\textcolor{blue}{\mathcal{H}} - \gamma \vS_C \otimes \vS_K]\big) ,
    \end{aligned}
    $%
}
\end{equation*}
and update a Kronecker-factored preconditioner $\vS$ without square-roots,
where \scalebox{0.78}{ $2\partial_{S^{-1}}\mathcal{L} \approx  \mathcal{H} - \gamma \vS = \mathcal{H}- \gamma \vS_C \otimes \vS_K $ } is due to Eq.~\ref{eq:grad_gauss}.
When $\mathcal{H}=\vH$, we obtain a root-free Shampoo-like update rule due to the simplification of the derivatives:

\vspace{-0.6cm}
\begin{equation*}
    \resizebox{0.85\hsize}{!}{%
    $
    \begin{aligned}
    2[ \textcolor{blue}{\partial_{S_C^{-1}} \mathcal{L}}]_{ij} &\approx \mathrm{Tr} \big( \frac{\partial \vS^{-1}}{\partial [\vS_C^{-1} ]_{ij}} [\vH - \gamma \vS_C \otimes \vS_K]\big) =  [ \vG \vS_K^{-1} \vG^\top - \gamma d \vS_C ]_{ij}\\
  2[\textcolor{blue}{\partial_{S_K^{-1}} \mathcal{L}} ]_{ij}& \approx \mathrm{Tr}\big( \frac{\partial \vS^{-1}}{\partial [\vS_K^{-1}]_{ij} } [\vH - \gamma \vS_C \otimes \vS_K] \big) = [ \vG^\top \vS_C^{-1} \vG - \gamma p\vS_K]_{ij}.
    \end{aligned}
    $%
}
\end{equation*}

\vspace{-0.4cm}
Notably, our root-free updates for $\vS_K$ and $\vS_C$ (c.f. Fig.~\ref{fig:matrix_methos_simple}) are correlated while Shampoo's updates for $\hat{\vS}_K$ and $\hat{\vS}_C$ are uncorrelated due to Shampoo's block-diagonal approximation of $\vH\approx {\hat{\vS}}_C^{\nicefrac{1}{2}} \otimes {\hat{\vS}}_K^{\nicefrac{1}{2}} $ and its moving average heuristics.
When the loss is scaled by averaging $B$ data points, we obtain a similar update using our empirical Fisher (c.f. Sec.~\ref{sec:second_order}) as a proper curvature approximation
$\mathcal{H}= \tilde{\vF}_{\text{scaled}}=B \vH$.

Our update is similar to other root-free second-order methods based on the multi-linear exponential family \citep{lin2019fast} or the tensor Gaussian distribution \citep{ren2021tensor} when using the outer product as a curvature approximation (i.e., $\mathcal{H}=\vH$).
Unlike these works, our update is both affine-invariant and scale-invariant of the loss and is applicable for other curvature approximations $\mathcal{H}$.

\paragraph{Structured inverse-free (IF) update}
To enable our update in half-precision,
we consider an IF update \citep{lin2023simplifying} by reparameterizing \scalebox{0.75}{$\vS^{-1}= \vS_C^{-1} \otimes \vS_K^{-1} =(\vC\vC^\top) \otimes (\vK\vK^\top) $} and  updating $\vC$ and $\vK$ instead of $\vS_C$ and $\vS_K$  thanks to the preconditioner invariance.
This update is IF and root-free (c.f. Fig.~\ref{fig:matrix_methos_simple}), avoiding numerically unstable matrix inversions and decompositions (c.f. Appx.~\ref{sec:app_matrix_inv_free} for a derivation).
Moreover, this update is equivalent to updating $\vS_C$ and $\vS_K$ up to  first-order accuracy (c.f. Appx.~\ref{app:invariance_ours_root_free}).
Last but not least, this update also allows the use of sparse factors $\vC$ and $\vK$ \citep{lin2023structured} to further lower memory consumption.
By using a sparse  $\vC$,
our approach can construct an expressive structured preconditioner $\vS_C$ or its inverse $\vS_C^{-1}$ that is difficult to achieve with other approaches.

We find that IF-Shampoo performs similarly to Shampoo regarding per-iteration progress on modern vision models such as vision transformers and mambas (see Fig.~\ref{fig:experiment-matrix-more}). However, our method can run in BFP-16, in contrast to Shampoo. We observe that one step of IF-Shampoo takes less than half the time of Shampoo (see Fig.~\ref{fig:experiment-matrix}) and requires less memory. This promising result will make matrix adaptive methods more prominent in modern large-scale training.

\section{Conclusion}
We investigated root-free adaptive methods and strengthened their motivation from a second-order perspective.
Surprisingly, we found empirically that removing the root not only \emph{closes} the generalization gap between adaptive methods and SGD on CNNs, but also \emph{maintains} the performance of root-based methods on vision attention models.
Removing the root eliminates the connection to sign descent which has been hypothesized to cause the gap on convolutional NNs and transformers. However, our findings highlight that adaptivity might be an important concept for the success of such methods that is currently overlooked, which poses new questions regarding the role of adaptivity and the understanding of adaptive methods.
Conceptually, we established the second-order view on root-free methods by viewing their gradient outer product as a novel empirical Fisher
that preserves scale-invariance of the loss and affine invariance.
Computationally,
we introduced the concept of preconditioner invariance to reparametrize preconditioners and developed matrix adaptive, inverse-free methods that stably work in BFP-16 and train faster than their root-based counterpart.
By using low-precision training, our IF-Shampoo has a similar iteration cost as diagonal methods while achieving better performance on modern NN models.

\section*{Acknowledgements}
We thank Emtiyaz Khan, Mark Schmidt,  Kirill Neklyudov, and Roger Grosse for helpful discussions at the early stage of this work.
Resources used in preparing this research were provided, in part, by the Province of Ontario, the Government of Canada through CIFAR, and companies sponsoring the Vector Institute.
Runa Eschenhagen is supported by ARM and the Cambridge Trust.
Richard E. Turner is supported by Google, Amazon, ARM, Improbable and EPSRC grant EP/T005386/1.

\section*{Impact Statement}
This paper introduces research aimed at improving deep learning optimization. While there are numerous potential societal implications stemming from our work, such as giving the ability to train neural networks more efficiently, we believe that none warrant specific emphasis within this context.

\bibliography{refs}
\bibliographystyle{icml2024}

\newpage
\appendix
\onecolumn
\section{Details on the Optimizer Updates}
\label{sec:app-details-optimizer-updates}
In \Cref{fig:rmsprop-full}, we show the algorithm for root-free RMSProp.
\Cref{fig:matrix_methos_full} summarizes the update rules of root-free Shampoo, as well as inverse- and root-free Shampoo, for one- and two-dimensional weight tensors.

\begin{figure*}[!t]
\center
	\fbox{
			\begin{minipage}{.5\textwidth}
             \textbf{Square-root-free RMSProp}

		\begin{algorithmic}[1]
               \STATE
            \footnotesize  Compute gradient $\vg\coloneqq\nabla  \ell_{\text{scaled}}(\vmu)$  \\
             \scalebox{0.9}{   $
            \vs  \leftarrow (1-\stepsize_2) \vs + \stepsize_2 {\color{red} B} \vg^2
             $} \\
            \STATE
             \scalebox{0.9}{   $
             \vm \leftarrow \alpha_1 \vm +  \vg/(\vs+\lambda)  + \kappa \vmu
             $} \\
            \STATE
          \scalebox{0.9}{    $
         \vmu \leftarrow \vmu  -\stepsize_1 \vm
 $}
				\end{algorithmic}
	\end{minipage}
	}
 \vspace{-0.1cm}   \caption{
 A full-fledged version of our square-root-free (diagonal) RMSProp, where $\alpha_1$, $\lambda$, $\kappa$ are the weight to include momentum, damping, weight decay, respectively.
 The preconditioner $\vs$ is initialized with $1$.
}\label{fig:rmsprop-full}
\end{figure*}

\begin{figure*}[!t]
\center
   \fbox{
			\begin{minipage}{.46\textwidth}
		\textbf{Square-root-free Shampoo (matrix case)}

				\begin{algorithmic}[1]
					            \STATE
            \footnotesize  Compute $\vG\coloneqq\mathrm{Mat}( \nabla   \ell_{\text{scaled}}(\vmu) )$  \\
             \scalebox{0.9}{   $
            \vS_C  \leftarrow (1-\stepsize_2 \gamma ) \vS_C + \frac{\stepsize_2 }{d} ( {\color{red} B } \vG \vS_K^{-1} \vG^\top + \lambda \mathrm{Tr}(\vS_K^{-1}) \vI_p)
             $} \\
             \scalebox{0.9}{   $
            \vS_K  \leftarrow (1-\stepsize_2 \gamma ) \vS_K + \frac{\stepsize_2 }{p} (  {\color{red} B } \vG^\top \vS_C^{-1} \vG + \lambda \mathrm{Tr}(\vS_C^{-1}) \vI_d )
             $} \\

            \STATE
          \scalebox{0.82}{    $
               \vM \leftarrow \alpha_1 \vM + \vS_C^{-1} \vG \vS_K^{-1} + \kappa  \mathrm{Mat}(\vmu)
          $}
            \STATE
          \scalebox{0.82}{    $
          \mathrm{Mat}(\vmu) \leftarrow    \mathrm{Mat}(\vmu) -  \stepsize_1  \vM
 $}

				\end{algorithmic}
	\end{minipage}
	}
   \fbox{
			\begin{minipage}{.52\textwidth}
		\textbf{Inverse-free Shampoo (matrix case)}

				\begin{algorithmic}[1]
            \STATE
            \footnotesize  Compute $\vG\coloneqq\mathrm{Mat}( \nabla  \ell_{\text{scaled}}(\vmu) )$  \\
             \scalebox{0.82}{   $
            \vm_C  \leftarrow  \alpha_2 \vm_C + \frac{(1-\alpha_2)}{2 d} ({\color{red} B }\vC^\top \vG \vK \vK^\top \vG^\top \vC + \lambda \mathrm{Tr}(\vK\vK^\top) \vC^\top \vC - d\gamma \vI_p  )
             $} \\
             \scalebox{0.82}{   $
            \vm_K  \leftarrow \alpha_2 \vm_K  + \frac{(1-\alpha_2)}{2 p}( {\color{red} B }\vK^\top \vG^\top \vC \vC^\top \vG \vK + \lambda \mathrm{Tr}(\vC\vC^\top)\vK^\top \vK -p\gamma \vI_d )
             $} \\
            \scalebox{0.82}{   $
            \vC  \leftarrow \vC \exp(-\stepsize_2  \vm_C  )
             $} %
             \\
            \scalebox{0.82}{   $
            \vK  \leftarrow \vK \exp(-\stepsize_2 \vm_K )
             $} %

               \STATE
          \scalebox{0.82}{    $
               \vM \leftarrow \alpha_1 \vM + \vC\vC^\top \vG \vK \vK^{T} +\kappa  \mathrm{Mat}(\vmu)
          $}
            \STATE
          \scalebox{0.82}{    $
          \mathrm{Mat}(\vmu) \leftarrow    \mathrm{Mat}(\vmu)   -\stepsize_1 \vM
 $}

				\end{algorithmic}
	\end{minipage}
	}
   \fbox{
			\begin{minipage}{.46\textwidth}
		\textbf{Square-root-free Shampoo (vector case)}

				\begin{algorithmic}[1]
					            \STATE
            \footnotesize  Compute $\vg\coloneqq \nabla   \ell_{\text{scaled}}(\vmu) $  \\
             \scalebox{0.9}{   $
            \vS  \leftarrow (1-\stepsize_2 \gamma ) \vS + \stepsize_2 ( {\color{red} B } \vg  \vg^\top + \lambda \vI)
             $} \\
            \STATE
          \scalebox{0.82}{    $
               \vm \leftarrow \alpha_1 \vm + \vS^{-1} \vg  + \kappa \vmu
          $}
            \STATE
          \scalebox{0.82}{    $
          \vmu \leftarrow   \vmu -  \stepsize_1  \vm
 $}

				\end{algorithmic}
	\end{minipage}
	}
   \fbox{
			\begin{minipage}{.46\textwidth}
		\textbf{Inverse-free Shampoo (vector case)}

				\begin{algorithmic}[1]
            \STATE
            \footnotesize  Compute $\vg\coloneqq \nabla  \ell_{\text{scaled}}(\vmu)$  \\
             \scalebox{0.82}{   $
            \vm_A  \leftarrow \alpha_2 \vm_A  + \frac{(1-\alpha_2)}{2 }( {\color{red} B }\vA^\top \vg \vg^\top \vA + \lambda \vA^\top \vA - \gamma \vI )
             $}
             \\
            \scalebox{0.82}{   $
            \vA  \leftarrow \vA \exp(-\stepsize_2 \vm_A )
             $} %

               \STATE
          \scalebox{0.82}{    $
               \vm \leftarrow \alpha_1 \vm + \vA \vA^{\top}\vg  +\kappa \vmu
          $}
            \STATE
          \scalebox{0.82}{    $
          \vmu \leftarrow    \vmu   -\stepsize_1 \vm
 $}

				\end{algorithmic}
	\end{minipage}
	}

 \vspace{-0.1cm}   \caption{
 A full-fledged version of our matrix methods, where $\alpha_1$, $\lambda$, $\kappa$ are the weight to include momentum, damping, weight decay, respectively.
 Note that we compute $\vm_C$ and $\vm_K$ using old $\vC$ and $\vK$. We also use old $\vS_C$ when updating $\vS_K$.
 We also include weight $\alpha_2$ for Riemannian momentum suggested by \citet{lin2023simplifying}.
 We approximate the matrix exponential $\exp(\vN)$ by its first-order truncation $\exp(\vN)\approx \vI + \vN$ as suggested by \citet{lin2023simplifying}.
 For our inverse-free Shampoo, we store and update preconditioner factors $\vC$, $\vm_C$, $\vK$, $\vm_K$ in half precision.
 For numerical stability in half precision, we update $\vC$ and $\vK$ as
$\vC \leftarrow \vC \big(\vI - \frac{\stepsize \vm_C}{ \max\{ \|\vm_C\|, 1 \}} \big) $ and
$\vK \leftarrow \vK \big(\vI - \frac{\stepsize  \vm_K}{ \max\{ \|\vm_K\|, 1 \}} \big) $, where $\|\cdot\|$ denotes the Frobenius norm.
Using  the matrix norm also allows us to make  discretization errors (e.g., the term $O(\stepsize_2^2) $ in Claim~\ref{claim:shampoo_invariance_free}) in the updates  negligible, where $\stepsize_2\coloneqq\frac{\stepsize }{ \max\{ \|\vm_C\|, 1 \}} $ and $\alpha_2=0$ is the case considered in the claim.
 We also show an update for vector-shaped parameters, which can be obtained from the matrix case by treating a vector as a matrix with a single column and freezing the pre-conditioner for the dimension of size 1 to be one.
  }\label{fig:matrix_methos_full}
\end{figure*}

\subsection{Inverse- and root-free Shampoo for $N$-dimensional tensors.}
\label{sec:app_inverse_and_rootfree_shampoo_Nd}

Here, we describe our inverse- and root-free Shampoo's update for $N$-dimensional tensors.
We use the tensor network notation from \citet{dangel2023convolutions}, which is easier to parse than the mathematical expressions in index notation.
For an $N$d weight tensor $\vW \in \mathbb{R}^{d_1 \times \dots \times d_N}$, the pre-conditioner is built of $N$ matrices $\vK_1, \dots, \vK_N$ and their momenta $\vm_1, \dots, \vm_N$ where $\vK_n \in \mathbb{R}^{d_n \times d_n}$ and $\vm_n \in \mathbb{R}^{d_n \times d_n}$. To perform a step:

{
  \def\K#1{%
    \tensornode{K#1}%
    {$\vK_{#1}$}%
    {{j#1}}%
    {{}}%
    {{i#1}}%
    {{}}%
  }%
  \def\KKT#1{%
    \tensornode{KKT#1}%
    {$\vK_{#1}\vK_{#1}^{\top}$}%
    {{j#1}}%
    {{}}%
    {{i#1}}%
    {{}}%
  }%
  \def\Kdots#1{%
    \tensornode{K#1}%
    {$\dots$}%
    {{j#1}}%
    {{}}%
    {{i#1}}%
    {{}}%
  }%

  \def\W{%
    \tensornode{W}%
    {\rule{0pt}{16ex}$\vW$\rule[-14ex]{0pt}{16ex}}%
    {{i1,i2,i3,in,i5,iN}}%
    {{}}%
    {{}}%
    {{}}%
  }%

  \def\M{%
    \tensornode{M}%
    {\rule{0pt}{16ex}$\vM$\rule[-14ex]{0pt}{16ex}}%
    {{i1,i2,i3,in,i5,iN}}%
    {{}}%
    {{}}%
    {{}}%
  }%

  \def\G{%
    \tensornode{G}%
    {\rule{0pt}{16ex}$\vG$\rule[-14ex]{0pt}{16ex}}%
    {{i1,i2,i3,in,i5,iN}}%
    {{}}%
    {{}}%
    {{}}%
  }%

  \def\Ghat{%
    \tensornode{Ghat}%
    {\rule{0pt}{16ex}$\hat{\vG}$\rule[-14ex]{0pt}{16ex}}%
    {{j1,j2,j3,jn,j5,jN}}%
    {{}}%
    {{}}%
    {{}}%
  }%

  \def\Ghatn{%
    \tensornode{Ghatn}%
    {$\hat{\vG}_n$}%
    {{k}}%
    {{}}%
    {{jn}}%
    {{}}%
  }%

  \newsavebox{\boxGhat}
  \begin{lrbox}{\boxGhat}
    \begin{tikzpicture}[ultra thick]
      \matrix[row sep=0ex,column sep=2ex]{%
        \Ghat
        &
        \node {\quad$=$};
        &
        \G
        \\
      };%
      \draw
      (Ghat-j1) node[index, right] {$j_1$}%
      (Ghat-j2) node[index, right] {$j_2$}%
      (Ghat-j3) node[index, right] {$\dots$}%
      (Ghat-jn) node[index, right] {$j_n$}%
      (Ghat-j5) node[index, right] {$\dots$}%
      (Ghat-jN) node[index, right] {$j_N$}%
      ;%
      \begin{scope}[shift={($(G-i1)+(7ex,0)$)}]
        \K{1}
        \draw %
        \contract{G}{K1}{i1}{$i_1$}{out=0, in=180};
      \end{scope}
      \begin{scope}[shift={($(G-i2)+(7ex,0)$)}]
        \K{2}
        \draw %
        \contract{G}{K2}{i2}{$i_2$}{out=0, in=180};
      \end{scope}
      \begin{scope}[shift={($(G-i3)+(7ex,0)$)}]
        \Kdots{3}
        \draw %
        \contract{G}{K3}{i3}{$\dots$}{out=0, in=180};
      \end{scope}
      \begin{scope}[shift={($(G-in)+(7ex,0)$)}]
        \K{n}
        \draw %
        \contract{G}{Kn}{in}{$i_n$}{out=0, in=180};
      \end{scope}
      \begin{scope}[shift={($(G-i5)+(7ex,0)$)}]
        \Kdots{5}
        \draw %
        \contract{G}{K5}{i5}{$\dots$}{out=0, in=180};
      \end{scope}
      \begin{scope}[shift={($(G-iN)+(7ex,0)$)}]
        \K{N}
        \draw %
        \contract{G}{KN}{iN}{$i_N$}{out=0, in=180};
      \end{scope}
      \draw
      (K1-j1) node[index, right] {$j_1$}%
      (K2-j2) node[index, right] {$j_2$}%
      (K3-j3) node[index, right] {$\dots$}%
      (Kn-jn) node[index, right] {$j_n$}%
      (K5-j5) node[index, right] {$\dots$}%
      (KN-jN) node[index, right] {$j_N$}%
      ;%
    \end{tikzpicture}
  \end{lrbox}

  \newsavebox{\boxGhatn}
  \begin{lrbox}{\boxGhatn}
    \begin{tikzpicture}[ultra thick]
      \matrix[row sep=0ex,column sep=2ex]{%
        \Ghatn
        &
        \node {\quad\qquad\qquad\qquad\qquad\qquad$=$};
        &
        \Ghat
        \\
      };%
      \draw
      (Ghatn-jn) node[index, left] {$j_n$}%
      (Ghatn-k) node[index, right] {$(j_1, j_2, \dots, j_{n-1}, j_{n+1}, \dots, j_N)$}%
      (Ghat-jn) node[index, right] {$j_n$}%
      ;%
      \coordinate (join) at ($(Ghat)+(13ex,0)$);%
      \draw[arrows=-{Triangle}] (join) to ++(0.01ex,0);%
      \draw (join) to ++(2ex,0) node [index, right] {$(j_1, j_2, \dots, j_{n-1}, j_{n+1}, \dots, j_N)$};%
      \draw (Ghat-j1) [out=0, in=180] to (join);%
      \draw (Ghat-j2) [out=0, in=180] to (join);%
      \draw (Ghat-j3) [out=0, in=180, dotted] to (join);%
      \draw (Ghat-j5) [out=0, in=180, dotted] to (join);%
      \draw (Ghat-jN) [out=0, in=180] to (join);%
    \end{tikzpicture}
  \end{lrbox}

  \newsavebox{\boxM}
  \begin{lrbox}{\boxM}
    \begin{tikzpicture}[ultra thick]
      \matrix[row sep=0ex,column sep=2ex]{%
        \M
        &
        \node {$\leftarrow$\quad$\alpha_1$};
        &
        \M
        &
        \node {$+$};
        &
        \G
        &
        \node {\qquad\qquad\quad\quad$+$ \quad$\kappa$};
        &
        \W
        \\
      };%
      \begin{scope}[shift={($(G-i1)+(7ex,0)$)}]
        \KKT{1}
        \draw (G-i1) to (KKT1-i1);
      \end{scope}
      \begin{scope}[shift={($(G-i2)+(7ex,0)$)}]
        \KKT{2}
        \draw (G-i2) to (KKT2-i2);
      \end{scope}
      \begin{scope}[shift={($(G-i3)+(7ex,0)$)}, tensor/.append style={minimum width = 7.35ex}]
        \Kdots{3}
        \draw (G-i3) to (K3-i3);
      \end{scope}
      \begin{scope}[shift={($(G-in)+(7ex,0)$)}]
        \KKT{n}
        \draw (G-in) to (KKTn-in);
      \end{scope}
      \begin{scope}[shift={($(G-i5)+(7ex,0)$)}, tensor/.append style={minimum width = 7.35ex}]
        \Kdots{5}
        \draw (G-i5) to (K5-i5);
      \end{scope}
      \begin{scope}[shift={($(G-iN)+(7ex,0)$)}]
        \KKT{N}
        \draw (G-iN) to (KKTN-iN);
      \end{scope}
    \end{tikzpicture}
  \end{lrbox}

  \begin{enumerate}
  \item Compute the gradient $\vG$ ($N$d tensor), then
    \begin{enumerate}
    \item Compute $\hat{\vG}$ ($N$d tensor) by multiplying each $\vK_n^{\top}$ onto the $n$-th axis of $\vG$:
      \begin{align}
        \text{
        \scalebox{1.0}{%
        \usebox{\boxGhat}
        }
        }
      \end{align}

    \item Compute the self-outer products $\{ \vK_n \vK_n^{\top} \}_{n=1}^N$, their traces $\{ \operatorname{Tr}(\vK_n \vK_n^{\top}) \}_{n=1}^N$, and define $D_{\setminus i} = \prod_{n \neq i}^N d_n$
    \end{enumerate}

  \item For all $n = 1, \dots, N$, update the $n$-th axis' pre-conditioner $\vK_n$ and its momentum $\vm_n$
    \begin{enumerate}
    \item Flatten all other axes of $\hat{\vG}$ to obtain the matrix $\hat{\vG}_n \in \mathbb{R}^{d_n \times D_{\setminus n}}$
      \begin{align}
        \scalebox{1.0}{%
        \usebox{\boxGhatn}
        }
      \end{align}

    \item Update the momentum $\vm_n$
      \begin{align}
        \vm_n \leftarrow \alpha_2 \vm_n + \frac{(1 - \alpha_2)}{D_{\setminus n}}
        \left\{
        B \hat{\vG}_{n} \hat{\vG}_{n}^{\top}
        + \lambda \left[ \prod_{i \neq n} \operatorname{Tr}(\vK_i \vK_i^{\top}) \right ] \vK_n \vK_n^{\top}
        - \gamma D_{\setminus n} \vI
        \right\}
      \end{align}

    \item Update the pre-conditioner matrix $\vK_n$, using a first-order truncation of the exponential map
      \begin{align}
        \vK_n \leftarrow \vK_n \exp \left( -\beta_2 \vm_n \right) \approx \vK_n - \beta_2 \vK_n \vm_n
      \end{align}
    \end{enumerate}

  \item Update the neural network parameters $\vW$ ($N$d tensor)
    \begin{enumerate}
    \item Compute the momentumized and weight-decayed update direction $\vM$ ($N$d tensor)
      \begin{align}
        \scalebox{1.0}{%
        \usebox{\boxM}
        }
      \end{align}

    \item Update the weights
      \begin{align}
        \vW \leftarrow \vW - \beta_1 \vM
      \end{align}
    \end{enumerate}

  \end{enumerate}
}

\section{Example: Affine Invariance of Root-Free Methods}
\label{sec:app-example-affine-invariance}
We demonstrate the affine invariance by an example and show how adding the root breaks the invariance.
Consider a loss function $l_a(a)=\half a^2$ with an initial point $a_0=2$ , a root-based update  is \scalebox{0.78}{ $a_{\text{new}}= a_0 -  s_a^{-1} g_a = 2 - 1 =1 $}, where gradient $g_a=\nabla_a l_a(a)=a_0=2$ and preconditioner \scalebox{0.78}{ $s_a= \sqrt{g_a^2} = |a_0|=2$}.
Now, consider a reparameterized loss as  $l_2(b)=\half (2b)^2$ with an initial point $b_0$, where $a=2b$. Thus, $b_0=1$ when $a_0=2$.
The update  becomes \scalebox{0.78}{$b_{\text{new}}= b_0 - s_b^{-1} g_b = 1 - 1 = 0 $}, where gradient $g_b=\nabla_b l_b(b)= 4b_0=4$ and preconditioner $s_b= \sqrt{g_b^2} = 4|b_0|=4$. Unfortunately, the updated $a_{\text{new}}=1 $ is not equivalent to the updated $ b_{\text{new}}=0$ since $a_{\text{new}} \neq 2 b_{\text{new}}$.

Now, consider a square-root-free update for the original loss as $a_{\text{new}}= a_0 -  s_a^{-1} g_a = 2 - 0.5 =1.5 $, where gradient $g_a=\nabla_a l_a(a)=a_0=2$ and preconditioner $s_a= g_a^2 = a_0^2=4$.
Similarly,  the update for the reparameterized loss is $b_{\text{new}}= b_0 - s_b^{-1} g_b = 1 - 0.25 =0.75 $, where gradient $g_b=\nabla_b l_b(b)=4b_0=4$ and preconditioner $s_b= g_b^2 = 16 b_0^2=16$.
Note that the updated $a_{\text{new}}=1.5$ is  equivalent to the updated $ b_{\text{new}}=0.75$ since $a_{\text{new}} = 2 b_{\text{new}}$. Thus, removing the root preserves the affine invariance.

\section{Proof of Claim ~\ref{claim:unbiasedness}}
\label{sec:app-proofunbiasedness}
We first show that our Fisher matrix coincides with the standard Fisher.
\begin{align}
    \vF_{\text{new}}(\vmu) = \vF_{\text{standard}}(\vmu)= \sum_{i=1}^{N} E_{y_i \sim p}\big[ \nabla_\mu \log p(y_i| \vx_i;\vmu) \nabla_\mu^\top \log p(y_i| \vx_i;\vmu) \big]
\end{align}
Similarly, we can show our mini-batch Fisher coincides with the standard mini-batch Fisher.
\begin{align}
    \vF_{\text{mini}}(\vmu) = \vF_{\text{standard-mini}}(\vmu)\coloneqq\sum_{i=1}^{B} E_{y_i \sim p}\big[ \nabla_\mu \log p(y_i| \vx_i;\vmu) \nabla_\mu^\top \log p(y_i| \vx_i;\vmu) \big]
\end{align}
Thus, it is easy to see that $\frac{1}{B}\vF_{\text{mini}}(\vmu)$ is an unbiased estimation of
 $\frac{1}{N}\vF_{\text{new}}(\vmu)$ since the standard mini-batch Fisher is an
 unbiased estimation of the standard full-batch Fisher.

 Now, we show that
 our Fisher matrix coincides with the standard Fisher.
 Recall that we define the joint distribution over labels is $p(\mathbf{y} | \vX;\vmu)= \prod_{i=1}^{N} p(y_i|\vx_i;\vmu)$
 \begin{align*}
    &\vF_{\text{new}}(\vmu) \\
    =& E_{\mathbf{y} \sim p}\big[ \nabla_\mu \log p(\mathbf{y} | \vX;\vmu) \nabla_\mu^\top \log p(\mathbf{y} | \vX;\vmu) \big] \\
    =& E_{\mathbf{y} \sim p}\big[ \sum_{i}\big( \nabla_\mu \log p(y_i | \vx_i;\vmu) \big) \big( \sum_j \nabla_\mu^\top \log p(y_j | \vx_j;\vmu)\big) \big] \\
    =&  E_{\mathbf{y} \sim p}\big[ \sum_{i = j} \big( \nabla_\mu \log p(y_i | \vx_i;\vmu) \big) \big( \nabla_\mu^\top \log p(y_j | \vx_j;\vmu)\big) \big] +  E_{\mathbf{y} \sim p}\big[ \sum_{i \neq j} \big( \nabla_\mu \log p(y_i | \vx_i;\vmu) \big) \big(  \nabla_\mu^\top \log p(y_j | \vx_j;\vmu)\big) \big] \\
    =&  \sum_{i } E_{y_i \sim p}\big[ \big( \nabla_\mu \log p(y_i | \vx_i;\vmu) \big) \big(  \nabla_\mu^\top \log p(y_i | \vx_i;\vmu)\big) \big] +  E_{\mathbf{y} \sim p}\big[ \sum_{i \neq j} \big( \nabla_\mu \log p(y_i | \vx_i;\vmu) \big) \big(  \nabla_\mu^\top \log p(y_j | \vx_j;\vmu)\big) \big] \\
    =&  \vF_{\text{standard}}(\vmu)+ \sum_{i \neq j} E_{\mathbf{y} \sim p}\big[  \big( \nabla_\mu \log p(y_i | \vx_i;\vmu) \big) \big(  \nabla_\mu^\top \log p(y_j | \vx_j;\vmu)\big) \big]  \\
     =&  \vF_{\text{standard}}(\vmu)
\end{align*} where the last line is due to
the independence and normalization of per-sample distributions  as shown below.

Since each per-sample distribution is independent, we have
\begin{align*}
    \sum_{i \neq j} E_{\mathbf{y} \sim p}\big[  \big( \nabla_\mu \log p(y_i | \vx_i;\vmu) \big) \big(  \nabla_\mu^\top \log p(y_j | \vx_j;\vmu)\big) \big] =
    \sum_{i \neq j} \underbrace{ E_{y_i \sim p}\big[  \nabla_\mu \log p(y_i | \vx_i;\vmu) \big]}_{=\mathbf{0}} E_{y_j \sim p}\big[  \nabla_\mu^\top \log p(y_j | \vx_j;\vmu)\big) \big] = \mathbf{0}
\end{align*} where we make use of the following result as the per-sample distribution is normalized.
\begin{align*}
    E_{y_i \sim p}\big[  \nabla_\mu \log p(y_i | \vx_i;\vmu) \big]
    = \int   \nabla_\mu  p(y_i | \vx_i;\vmu) dy_i = \nabla_\mu \underbrace{ \int    p(y_i | \vx_i;\vmu) dy_i}_{=1} = \mathbf{0}.
\end{align*}

\section{Proof of Claim ~\ref{claim:affine_inv}}
\label{sec:app-proofaffine-invariance}
Recall that
the (unscaled) optimization problem in \eqref{eq:optobj} is
\begin{align}
    \min_{\mu} \ell(\vmu)
\end{align}
Now, consider reparametrizing $\vmu$ with a known non-singular matrix $\vA$ and a constant vector $\vc$ as  $\vmu = \vA \vm + \vc$.
In this case, the optimization problem becomes
\begin{align}
 \min_{m} \ell^{\text{rep}}(\vm ) \coloneqq   \ell(\vA \vm+\vc)
\end{align}
We will show that a square-root-free method is affine invariant at each step.
In other words, if we use the same square-root-free method to solve these two problems, they are equivalent.

For the first problem,
 the method takes the following step at iteration $t$
\begin{align}
  \vS_{t+1} = (1-\stepsize_2 \gamma )\vS_{t} + \stepsize_2 \vH_t, \,\,\,\, \vmu_{t+1} = \vmu_t - \stepsize_1  \vS_{t+1}^{-1} \vg_t
\end{align} where $\vg_t \coloneqq  \nabla_\mu \ell(\vmu) \big|_{\mu= \mu_t } $ and $\vH_t = \vg_t \vg_t^\top $.

For the second problem, we assume $\vS_0^{\text{rep}}$ is initialized with $ \vA^\top \vS_0 \vA  $ and $ \vA^{-1} (\vmu_0-\vc) = \vm_0 $ since $\vA $ and $\vc$ are known.
In this case, the square-root-free update at the first iteration becomes
\begin{align*}
      \vS_{1}^{\text{rep}} &= (1-\stepsize_2 \gamma )\vS_{0}^{\text{rep}} + \stepsize_2 \vH_0^{\text{rep}} =   (1-\stepsize_2 \gamma )\vS_{0}^{\text{rep}} +  \stepsize_2  \vA^\top  \vH_0\vA = \vA^\top \underbrace{ \big (  (1-\stepsize_2 \gamma ) \vS_0 +  \stepsize_2     \vH_0  \big )}_{ = \vS_1 } \vA \\
      \vm_{1} &= \vm_0 - \stepsize_1  \big( \vS_{1}^{\text{rep}} \big)^{-1} \vg_0^{\text{rep}} = \vm_0 - \stepsize_1  \vA^{-1} \vS_1^{-1} \vA^{-T} \vA^\top \vg_0 = \vA^{-1} (\vmu_0 -\vc) - \stepsize_1  \vA^{-1} \vS_1^{-1}\vg_0  \\
      & =   \vA^{-1} ( \vmu_0 -\vc - \stepsize_1 \vS_1^{-1}\vg_0  ) = \vA^{-1} (\vmu_1-\vc)
\end{align*} where we use the following identities when $t=0$.
\begin{align*}
    \vg_t^{\text{rep}} & \coloneqq  \nabla_m \ell^{\text{rep}}(\vm) \big|_{m= m_t } = \frac{\partial \vmu}{\partial \vm}|_{m= m_t } \nabla_\mu \ell^{\text{rep}}(\vm) \big|_{\mu= \mu_t } =  \vA^\top \nabla_\mu \ell(\vmu)|_{\mu= \mu_t } = \vA^\top \vg_t \\
    \vH_t^{\text{rep}} & \coloneqq  \vg_t^{\text{rep}} \big(  \vg_t^{\text{rep}} \big)^\top = \vA^\top  \vH_t\vA
\end{align*}
From above expressions, we can see that  both updates are equivalent at the first iteration since $\vmu_1 = \vA_1 \vm_1 + \vc$.
Similarly, we can show that both updates are equivalent at every iteration by induction.

Thus, we can see that full-matrix square-root-free method is affine invariance. For a diagonal square-root-free method, it only preserves a reparametrization invariance with $\vA$ being diagonal.
Likewise, we can show that the update is affine invariance in a scaled case when using our empirical Fisher.

\section{Derivation of Our Inverse-free and Square-root-free Full-AdaGrad}
\label{sec:app_inverse_free_adagrad}
We show that we can make adaptive methods inverse-free via re-parameterization.
This is useful to enables matrix methods to work in low-precision settings.

We consider performing NGD  on another parametrization $\veta\coloneqq(\vmu,\vS^{-1})$ of the Gaussian, where $\vS^{-1} \succ 0$ is known as the covariance matrix and is also positive-definite.
We can perform NGD under this parameterization as
\begin{equation*}
    \begin{aligned}
\begin{bmatrix}
 \vmu \\ \vS^{-1}
\end{bmatrix}
& \leftarrow
\begin{bmatrix}
 \vmu \\ \vS^{-1}
\end{bmatrix} - \stepsize_1
\begin{bmatrix}
\vS & \mathbf{0}\\
\mathbf{0} & -\half \frac{\partial \vS }{\partial \vS^{-1} }
\end{bmatrix}^{-1}\begin{bmatrix}
 \partial_\mu \mathcal{L}  \\  \partial_{S^{-1}} \mathcal{L}
\end{bmatrix}   = \begin{bmatrix}
\vmu -\stepsize_1 \vS^{-1} \partial_\mu \mathcal{L} \\
\vS^{-1} + 2 \stepsize_1 \partial_{S} \mathcal{L}
   \end{bmatrix} = \begin{bmatrix}
\vmu -\stepsize_1 \vS^{-1} \partial_\mu \mathcal{L} \\
\vS^{-1} - 2 \stepsize_1 \vS^{-1} [\partial_{S^{-1}} \mathcal{L}]\vS^{-1}
   \end{bmatrix},
    \end{aligned}
\end{equation*}
where \scalebox{0.78}{$\vF_{\text{gauss}}(\veta)=\begin{bmatrix}
\vS & \mathbf{0}\\
\mathbf{0} & -\half \frac{\partial \vS }{\partial \vS^{-1} }
\end{bmatrix}$ } is the FIM of the Gaussian and is block-diagonal under this parameterization $\veta$, and we use this gradient identity from matrix calculus \scalebox{0.78}{$\partial_S \mathcal{L} = -\vS^{-1} [ \partial_{S^{-1}} \mathcal{L} ] \vS^{-1}$} in the last step.

In the above update, we use the old
$\vS^{-1}$ as a preconditioner.
We can use the newly updated $\vS_{\text{new}}^{-1}$ as a preconditioner as shown in the follwoing update rule
\begin{equation} \label{eq:inv_update_mu}
  \vS_{\text{new}}^{-1} \leftarrow (1+\stepsize_1 \gamma) \vS^{-1} -  \stepsize_1 \vS^{-1}\mathcal{H}\vS^{-1},\,\, \vmu \leftarrow \vmu - \stepsize_1 \vS_{\text{new}}^{-1}  \nabla_\mu \ell(\vmu),
\end{equation} when performing NGD on parameterization \scalebox{0.78}{$(\vS\vmu, \vS^{-1})$} where $2\partial_{S^{-1}} \mathcal{L}=\mathcal{H} - \gamma \vS $ and curvature matrix $\mathcal{H}=\nabla_\mu^2 \ell(\vmu)$

This above update is inverse-free since we directly update $\vS^{-1}$ without inverting $\vS$.
To obtain inverse-free adaptive methods, we use the second moment as a Hessian approximation (e.g., $\mathcal{H} = \vH$) and introduce an additional learning rate $\stepsize_2$ to compensate for this approximation.

However, the update of $\vS^{-1}$ in \eqref{eq:inv_update_mu} does not guarantee that $\vS_{\text{new}}^{-1}$ is positive-definite even when $\mathcal{H}=\vH=\vg\vg^\top$ is semi-positive-definite.
Inspired by \citet{lin2020handling},
we propose to add a correction term highlighted in red in the update of $\vS^{-1}$ to satisfy the constraint as
\begin{equation}  \label{eq:inv_update_app}
  \vS^{-1}_{\text{new}} \leftarrow (1+\stepsize_2 \gamma) \vS^{-1} -  \stepsize_2 \vS^{-1}\vH \vS^{-1} + {\color{red} \frac{\stepsize_2^2}{2} \vW },
\end{equation} where
\scalebox{0.78}{
$\vW\coloneqq\vS^{-1} \vH \vS^{-1} \vH \vS^{-1} - 2 \gamma \vS^{-1} \vH\vS^{-1} + \gamma^2 \vS^{-1}$}.

By setting $\gamma=0$, we obtain an inverse-free update for square-root-free full-AdaGrad in Eq.~\eqref{eq:inv_update} as
\begin{equation*}
  \vS_{\text{new}}^{-1} \leftarrow  \vS^{-1} -  \stepsize_2 \vS^{-1}\vH \vS^{-1} + \frac{\stepsize_2^2}{2} \vS^{-1} \vH \vS^{-1} \vH \vS^{-1},\,\, \vmu \leftarrow \vmu - \stepsize_1 \vS_{\text{new}}^{-1} \vg.
\end{equation*} %

\begin{claim}
\label{claim:inv_free_root_free_adagrad}
The update in Eq.~\eqref{eq:inv_update} is guaranteed to be positive-definite when the current $\vS^{-1}$ is positive-definite.
\end{claim}

\begin{proof}
We prove that the update in \eqref{eq:inv_update_app} is
positive-definite. Therefore, the update in
Eq.~\eqref{eq:inv_update} is also positive-definite by setting $\gamma=0$.

When the current/initial $\vS^{-1}$ is positive-definite,
the updated $\vS_{\text{new}}^{-1}$ in \eqref{eq:inv_update_app} is guaranteed to be positive-definite
since
\begin{equation} \label{eq:cholsky_spd}
    \begin{aligned}
  (1+\stepsize_2 \gamma) \vS^{-1} -  \stepsize_2 \vS^{-1}\vH \vS^{-1} +  \frac{\stepsize_2^2}{2} \vW  =
 \vA (\vI -  \stepsize_2 \vN  + \frac{\stepsize_2^2}{2} \vN^2   ) \vA^\top
=  \half \vA \big[ \vI + (\vI - \stepsize_2 \vN )  (\vI - \stepsize_2 \vN )^\top \big] \vA^\top \succ 0,
    \end{aligned}
\end{equation} where  the current $\vS ^{-1}$ can be decomposed as $\vS^{-1}=\vA\vA^\top$, $\vA$ is a square non-singular matrix, and $\vN\coloneqq\vA^\top \vH \vA - \gamma \vI$ is a symmetric matrix.
\end{proof}

%

\section{Preconditioner Invariance for Full-matrix Square-root-free Adaptive Methods}
\label{sec:app_matrix_invar_full}
We will show that following claim to explicitly demonstrate the preconditioner invariance.
This invariance allows us to reparametrize a  preconditioner $\vS$ to obtain an inverse-free update scheme.
In other words, the updates of $\bar{\vS}^{-1}$ and $\vS$ are equivalent up to  first-order accuracy thanks to this invariance.

\paragraph{Claim  \ref{claim:inv_free_full_adagrad}}
Let  $\bar{\vS}^{-1}$ be the inverse of a preconditioner  updated according to the inverse-free scheme (Eq.~\eqref{eq:inv_update})
 with this initialization $\bar{\vS}_0^{-1} = \vS_0^{-1}$.
 If  $\bar{\vS}$ and $\vS$ are updated by using the same sequence of curvature approximations $\vH$, then
   $\bar{\vS}$ has  first-order accuracy of the root-free full-AdaGrad update of ${\vS}$ (Eq.~\eqref{eq:rf_adamat})  at each iteration, i.e., \( \bar{\vS}   =    \vS  + O(\stepsize_2^2)\).

\begin{proof}
In our proof, we do not exploit that $\mathcal{H}=\vH=\vg\vg^T$ has a low-rank structure to use the Woodbury matrix identity. Thus, our proof can be applied to cases when $\mathcal{H}\neq \vg\vg^T$.
We will show an equivalent relationship as  $\bar{\vS}_t^{-1} \vS_t = \vI + O(\stepsize_2^2)$ by induction. By the initialization, we know that this relationship holds when $t=0$.
Suppose that this relationship holds when $t=k-1$ so that
$\bar{\vS}_{k-1}^{-1} \vS_{k-1} = \vI + O(\stepsize_2^2)$.

Now, we show that
$\bar{\vS}_{k}^{-1} \vS_{k} = \vI + O(\stepsize_2^2)$.
According to \eqref{eq:inv_update}, $\bar{\vS}^{-1}$ is updated as
\begin{align*}
    \bar{\vS}^{-1}_k = \bar{\vS}^{-1}_{k-1} - \stepsize_2  \bar{\vS}^{-1}_{k-1} \vH_{k-1} \bar{\vS}^{-1}_{k-1} + \frac{\stepsize_2^2}{2}\bar{\vS}^{-1}_{k-1} \vH_{k-1}\bar{\vS}^{-1}_{k-1} \vH_{k-1} \bar{\vS}^{-1}_{k-1}
\end{align*}

According to
Eq.~\eqref{eq:rf_adamat},  the root-free full-AdaGrad update of $\vS$ is
\begin{align*}
\vS_k = \vS_{k-1} + \stepsize_2 \vH_{k-1}.
\end{align*} %

Thus, we have
\begin{align*}
    \bar{\vS}^{-1}_k {\vS}_k &= ( \bar{\vS}^{-1}_{k-1} - \stepsize_2  \bar{\vS}^{-1}_{k-1} \vH_{k-1} \bar{\vS}^{-1}_{k-1} + \frac{\stepsize_2^2}{2}\bar{\vS}^{-1}_{k-1} \vH_{k-1}\bar{\vS}^{-1}_{k-1} \vH_{k-1} \bar{\vS}^{-1}_{k-1}
) ( \vS_{k-1} + \stepsize_2 \vH_{k-1} ) \\
&= [ \bar{\vS}^{-1}_{k-1} - \stepsize_2  \bar{\vS}^{-1}_{k-1} \vH_{k-1} \bar{\vS}^{-1}_{k-1} + O(\stepsize_2^2)
] ( \vS_{k-1} + \stepsize_2 \vH_{k-1} ) \\
&=  \underbrace{\bar{\vS}^{-1}_{k-1} \vS_{k-1}}_{\vI + O(\stepsize_2^2)} - \stepsize_2  \bar{\vS}^{-1}_{k-1} \vH_{k-1} \underbrace{\bar{\vS}^{-1}_{k-1} \vS_{k-1}}_{\vI + O(\stepsize_2^2)} + \stepsize_2 \bar{\vS}^{-1}_{k-1} \vH_{k-1} + O(\stepsize_2^2) \\
&= \vI - \stepsize_2 \bar{\vS}^{-1}_{k-1} \vH_{k-1} + \stepsize_2  \bar{\vS}^{-1}_{k-1} \vH_{k-1} + O(\stepsize_2^2) \\
&= \vI + O(\stepsize_2)
\end{align*}
Thus, by induction, we have
$\bar{\vS}_{t}^{-1} \vS_{t} = \vI + O(\stepsize_2^2)$.
\end{proof}

Now, we discuss the case of RF-AdaGrad where we can compute the exact inversion due to the Woodbury matrix identity.
By the Woodbury matrix identity, we have
\begin{align*}
 \text{(RF-AdaGrad)} \,\,\,\,  \vS_k^{-1} & \leftarrow \big(\vS_{k-1} + \stepsize_2 \vg_{k-1}\vg_{k-1}^T\big)^{-1} \\
    &= \vS_{k-1}^{-1} - \frac{\stepsize_2}{1+\stepsize_2 \vg_{k-1}^T \vS_{k-1}^{-1} \vg_{k-1} } \vS_{k-1}^{-1} \vg_{k-1}\vg_{k-1}^T \vS_{k-1}^{-1} \\
    &= \vS_{k-1}^{-1} - \stepsize_2
    \vS_{k-1}^{-1} \vg_{k-1}\vg_{k-1}^T \vS_{k-1}^{-1} + \stepsize_2^2 \frac{\vg_{k-1}^T \vS_{k-1}^{-1} \vg_{k-1}}{1+\stepsize_2 \vg_{k-1}^T \vS_{k-1}^{-1} \vg_{k-1} }\vS_{k-1}^{-1} \vg_{k-1}\vg_{k-1}^T \vS_{k-1}^{-1} \\
    &= \vS_{k-1}^{-1} - \stepsize_2
    \vS_{k-1}^{-1} \vg_{k-1}\vg_{k-1}^T \vS_{k-1}^{-1} + O(\stepsize_2^2) \,\,\,\, \text{(Ours)}
\end{align*} where the last term is $O(\stepsize_2^2)$ since $1+\stepsize_2 \vg_{k-1}^T \vS_{k-1}^{-1} \vg_{k-1} \rightarrow 1$ when $\stepsize_2 \rightarrow 0$.
Thus, we can see that the RF-AdaGrad update scheme with the exact inverse and our inverse-free update scheme are equivalent up to first-order accuracy.

Now, we consider another inverse-free update scheme.
As shown in \citet{lin2023simplifying}, the update of $\vA$ is
 \begin{align}
 \label{eq:exp_inv_update}
     \vA = \vA \mathrm{exp}(-\frac{\stepsize_2}{2} (\vA^\top \vH \vA -\gamma \vI) )
 \end{align}  where $(\vA\vA^\top)^{-1}$ is used as a preconditioner.

\begin{claim}
Let $(\vA\vA^\top)^{-1}$ be a preconditioner, where $\vA$ is
 initialized so that $(\vA_0\vA_0^\top)^{-1} = \vS_0$.
 If ${\vA}$ is updated according to the inverse-free scheme (Eq.~\eqref{eq:exp_inv_update} with $\gamma=0$), and $\vA$ and $\vS$ are updated by using the same sequence of curvature approximations $\vH$,
  then  $(\vA\vA^\top)^{-1}$ has  first-order accuracy of the root-free full-AdaGrad update of ${\vS}$ (Eq.~\eqref{eq:rf_adamat})  at each iteration, i.e., \( (\vA\vA^\top)^{-1} =    \vS   + O(\stepsize_2^2)\).
\end{claim}

\begin{proof}
 We will show an equivalent relationship as  $\vA_t\vA_t^\top \vS_t = \vI + O(\stepsize_2^2)$ by induction. By the initialization, we know that this relationship holds when $t=0$.
Suppose that this relationship holds when $t=k-1$ so that
$\vA_{k-1}\vA_{k-1}^\top \vS_{k-1} = \vI + O(\stepsize_2^2)$.

Now, we show that
$\vA_{k}\vA_{k}^\top \vS_{k}  = \vI + O(\stepsize_2^2)$.
According to the above update, $\vA$ is updated as
 \begin{align}
     \vA_k = \vA_{k-1} \mathrm{exp}(-\frac{\stepsize_2}{2} \vA_{k-1}^\top \vH_{k-1} \vA_{k-1}  )
 \end{align} and the product is
 \begin{align*}
  \vA_k \vA_k^\top &= \vA_{k-1} \mathrm{exp}\big(-\stepsize_2 \vA_{k-1}^\top \vH_{k-1} \vA_{k-1}  \big) \vA_{k-1}^\top \\
  &= \vA_{k-1} \big( \vI  - \stepsize_2 \vA_{k-1}^\top \vH_{k-1} \vA_{k-1}  + O(\stepsize_2^2) \big) \vA_{k-1}^\top \\
  &=  \vA_{k-1}  \vA_{k-1}^\top - \stepsize_2  \vA_{k-1}\vA_{k-1}^\top \vH_{k-1} \vA_{k-1}\vC_{k-1}^\top + O(\stepsize_2^2)
 \end{align*}
Thus, we have
 \begin{align*}
  \vA_k \vA_k^\top \vS_k &=
  \big[ \vA_{k-1}  \vA_{k-1}^\top - \stepsize_2  \vA_{k-1}\vA_{k-1}^\top \vH_{k-1} \vA_{k-1}\vA_{k-1}^\top + O(\stepsize_2^2) \big] \big( \vS_{k-1} + \stepsize_2 \vH_{k-1} \big) \\
  &=  \underbrace{\vA_{k-1}  \vA_{k-1}^\top \vS_{k-1}}_{\vI + O(\stepsize_2^2)} - \stepsize_2  \vA_{k-1}\vA_{k-1}^\top \vH_{k-1} \underbrace{ \vA_{k-1}\vA_{k-1}^\top \vS_{k-1}}_{\vI + O(\stepsize_2^2)}  + \stepsize_2 \vA_{k-1}  \vA_{k-1}^\top \vH_{k-1} + O(\stepsize_2^2) \\
  &=\vI + O(\stepsize_2^2)
 \end{align*}
 Thus, by induction, we have
$\vA_t\vA_t^\top \vS_{t} = \vI + O(\stepsize_2^2)$.
\end{proof}

\section{ Scale-invariance and Affine-invariance of Our Inverse-free Updates  }
\label{app:invariance_full_root_free}

\begin{claim}
Our root-free  update shown in %
Eq.~\eqref{eq:inv_update}
is invariant to a scale transformation of the loss.
\end{claim}

\begin{proof}
Recall that
the (unscaled) optimization problem in \eqref{eq:optobj} is
\begin{align}
    \min_{\mu} \ell(\vmu)
\end{align}

The scaled problem is
\begin{align}
 \min_{\bar{\mu}} \ell_{\text{scaled}}(\bar{\vmu} ) \coloneqq  \frac{1}{B} \ell(\bar{\vmu})
\end{align}

We will show that
our update is invariant at each step.
In other words, $\vmu=\bar{\vmu}$ holds for every iteration.
We assume $\vmu=\bar{\vmu}$ holds at the initialization step.

Recall that
our update rule  (Eq.~\eqref{eq:inv_update}) to solve the original problem is
\begin{align*}
  \vS^{-1} &=  \vS^{-1} -  \stepsize_2 \vS^{-1} \tilde{\vF}_{\text{new}} \vS^{-1} +  \frac{\stepsize_2^2}{2} \vS^{-1} \tilde{\vF}_{\text{new}}  \vS^{-1} \tilde{\vF}_{\text{new}} \vS^{-1} \\
    \vmu & = \vmu - \stepsize_1  \vS^{-1} \vg
\end{align*} where $\vg\coloneqq\nabla_\mu \ell(\vmu)$ and  $\tilde{\vF}_{\text{new}}= \vH = \vg\vg^\top$. Recall that the gradient outer product $\vH$ coincides with our empirical Fisher $\tilde{\vF}_{\text{new}}$ in this case.

In the scaled case, our update  is
\begin{align*}
  \bar{\vS}^{-1} & =  \bar{\vS}^{-1} -  \stepsize_2 \bar{\vS}^{-1} \tilde{\vF}_{\text{scaled}} \bar{\vS}^{-1} +  \frac{\stepsize_2^2}{2} \bar{\vS}^{-1} \tilde{\vF}_{\text{scaled}} \bar{\vS}^{-1}  \tilde{\vF}_{\text{scaled}} \bar{\vS}^{-1} \\
    \bar{\vmu} & = \bar{\vmu} - \stepsize_1  \bar{\vS}^{-1} \bar{\vg}
\end{align*}
 where $\bar{\vg}\coloneqq\nabla_{\bar{\mu}} \ell_{\text{scaled}}(\bar{\vmu})=\frac{1}{B}\vg$ and
 $\tilde{\vF}_{\text{new}}= B \bar{\vH} = B \bar{ \vg}\bar{\vg}^\top= \frac{1}{B}\vH$

We first assume that  $\bar{\vS}^{-1}= B \vS^{-1} $.
Therefore, we can  show that $\vmu = \bar{\vmu}$ since their descent directions are the same as
$
\bar{\vS}^{-1} \bar{\vg} = \big(B \vS^{-1} \big) \big( \frac{1}{B} \vg\big) = \vS^{-1} \vg
$.

Now, we show that  $\bar{\vS}= \frac{\vS}{B} $ at every iteration. We assume $\bar{\vS}$ is initialized so that this relationship holds in the base case.
Note that
we have the following relationship
\begin{align*}
    \bar{\vS}_{\text{new}}^{-1}  &= \underbrace{
   \bar{\vS}^{-1}}_{B S^{-1}} -  \stepsize_2  \bar{\vS}^{-1} \tilde{\vF}_{\text{scaled}} \bar{\vS}^{-1} +   \frac{\stepsize_2^2}{2} \bar{\vS}^{-1} \tilde{\vF}_{\text{scaled}} \bar{\vS}^{-1}  \tilde{\vF}_{\text{scaled}} \bar{\vS}^{-1} \\
   &= B\vS^{-1} - \stepsize_2  \big(B\vS^{-1}\big) \frac{\vH}{B}  \big(B\vS^{-1}\big)   +\frac{\stepsize_2^2}{2}  \big(B\vS^{-1}\big)  \frac{\vH}{B} \big(B\vS^{-1}\big)   \frac{\vH}{B}  \big(B\vS^{-1}\big)  \\
   &= B \big( \vS^{-1} -  \stepsize_2 \vS^{-1} \vH \vS^{-1} +  \frac{\stepsize_2^2}{2} \vS^{-1} \vH  \vS^{-1}  \vH \vS^{-1} \big) \\
   &= B \vS_{\text{new}}^{-1}
\end{align*}
Thus, we can use induction to establish this relationship.

\end{proof}

\begin{claim}
Our root-free  update  in Eq.~\eqref{eq:inv_update} is  invariant up to an  affine transformation.
\end{claim}

\begin{proof}
This proof is similar to the proof in Appx.\ref{sec:app-proofaffine-invariance}.
Recall that
the (unscaled) optimization problem in \eqref{eq:optobj} is
\begin{align}
    \min_{\mu} \ell(\vmu)
\end{align}
For simplicity, we only consider the linear transformation without a translation.
Now, consider reparametrizing $\vmu$ with a known non-singular matrix $\vA$  as  $\vmu = \vA \bar{\vmu}$ .
In this case, the optimization problem becomes
\begin{align}
 \min_{\bar{\mu}} \ell^{\text{rep}}(\bar{\vmu} ) \coloneqq   \ell(\vA \bar{\mu})
\end{align}

Let $\vS$ and $\bar{\vS}$ be preconditioners for the first problem and the second problem, respectively.
We will show that
$\bar{\vS}^{-1}= \vA^{-1} \vS^{-1} \vA^{-T}$.
Once this relationship is established, we can use a similar proof technique in Appx.~\ref{sec:app-proofaffine-invariance}  to complete the proof.

We can establish this relationship by induction.
Note that this relationship holds in the base case thanks to our initialization.

For the first problem,
 $\vS^{-1}$ is updated  as
\begin{align*}
  \vS^{-1}_{\text{new}} &=  \vS^{-1} -  \stepsize_2 \vS^{-1} \vH \vS^{-1} +  \frac{\stepsize_2^2}{2} \vS^{-1} \vH  \vS^{-1} \vH \vS^{-1}
\end{align*}

By induction, we know that $\bar{\vS}^{-1}= \vA^{-1} \vS^{-1} \vA^{-T}$ at the current iteration.
At a new iteration,
the preconditioner $\bar{\vS}^{-1}$ for the reparametrized problem is updated as
\begin{align*}
  \bar{\vS}^{-1}_{\text{new}} &= \underbrace{ \bar{\vS}^{-1}}_{  \vA^{-1} \vS^{-1} \vA^{-T} } -  \stepsize_2 \bar{\vS}^{-1} \underbrace{\bar{\vH} }_{ \vA^\top  \vH\vA} \bar{\vS}^{-1} +  \frac{\stepsize_2^2}{2} \bar{\vS}^{-1} \bar{\vH}  \bar{\vS}^{-1} \bar{\vH} \bar{\vS}^{-1} \\
  &=  \vA^{-1} \vS^{-1} \vA^{-T} - \stepsize_2  \vA^{-1} \vS^{-1} \vH \vS^{-1} \vA^{-T} +
   \frac{\stepsize_2^2}{2} \vA^{-1} {\vS}^{-1} {\vH}  {\vS}^{-1} {\vH} {\vS}^{-1} \vA^{-T} \\
   &= \vA^{-1} \big[  \vS^{-1} -  \stepsize_2 \vS^{-1} \vH \vS^{-1} +  \frac{\stepsize_2^2}{2} \vS^{-1} \vH  \vS^{-1} \vH \vS^{-1} \big] \vA^{-T} \\
   &= \vA^{-1} \vS^{-1}_{\text{new}}\vA^{-T}
\end{align*}
where we use the following identities.
\begin{align*}
    \bar{\vg} & \coloneqq  \nabla_{\bar{\mu}} \ell^{\text{rep}}(\bar{\vmu}) = \vA^\top \vg \\
    \bar{\vH} & \coloneqq  \bar{\vg}  \bar{\vg}^\top = \vA^\top  \vH\vA
\end{align*}
Thus, by induction, this relationship holds for every iteration.

\end{proof}

We can similarly show that another inverse-free update (Eq.~\eqref{eq:exp_inv_update}) is both scale and affine invariant.

\section{Derivation of Our Kronecker-factored Inverse-free Methods}
\label{sec:app_matrix_inv_free}
We consider reparametrize the inverse preconditioner (covariance) as $\vS^{-1}=(\vC\vC^\top) \otimes (\vK \vK^\top)$.  To obtain an inverse-free update scheme, we consider directly update $\vC$ and $\vK$ to bypass the need for matrix inversion, where $\vC \in \mathcal{R}^{p \times p } $ and $\vK \in \mathcal{R}^{d \times d} $ are square non-singular matrices.
Recall that we consider the curvature approximation $\mathcal{H}=\vH=\vg\vg^\top $ as a term of the partial derivative as $2\partial_{S^{-1}}\mathcal{L} \approx \vH - \gamma \vS   =  \vg\vg^\top - \gamma (\vC\vC^\top)^{-1} \otimes (\vK \vK^\top)^{-1}  $.
We consider the local coordinates proposed by \citet{lin2023simplifying}.  For simplicity, we only update the block $\vC$ while keeping blocks $\vmu$ and $\vK$ frozen.
Indeed, the blocks can be updated simultaneously.
To update the block $\vC$ at iteration $t$, the local coordinate $\veta_C$ associated to $\vC_t$  is defined as $\vS^{-1} =(\vC \vC^\top) \otimes (\vK \vK^\top) $, where $\vC = \vC_t \mathrm{Exp}( \frac{ \veta_C}{ \sqrt{2 d} } )$.   $\veta_C$ is a square symmetric matrix and it can be singular.
\citet{lin2023simplifying} show that the block approximated FIM of the Gaussian can be orthonormal at the origin in this local coordinate system as $\vF_{\text{gauss}}( \veta_C^\text{cur})   = \vI$. Importantly, the origin in this system represents the current block $\vC_t$  as $ \vC_t \equiv \vC_t \mathrm{Exp}( \frac{ \veta_C^\text{cur}}{ \sqrt{2 d} } ) $ where $\veta_C^\text{cur} = \mathbf{0}$. Thus, we can perform NGD in this local coordinate system as
\begin{align}
    \veta_C^{\text{new}} & \leftarrow \veta_C^\text{cur} - \stepsize_2 \left( \vF_{\text{gauss}}( \veta_C^\text{cur}) \right)^{-1} \partial_{\eta_C } \mathcal{L}   = \mathbf{0} - \stepsize_2 \partial_{\eta_C } \mathcal{L}  \\
    \vC_{t+1} & \leftarrow \vC_t \mathrm{Exp}( \frac{ \veta_C^{\text{new}} }{ \sqrt{2 d} } ) \approx \vC_t ( \vI + \frac{ \veta_C^{\text{new}} }{ \sqrt{2 d} } )
\end{align}
where we can use the chain rule to compute the partial derivative at $\veta_C^\text{cur}=\mathbf{0}$ and $\mathcal{H}=\vH$
\begin{align}
    2\partial_{\eta_C} \mathcal{L}  =2 \frac{ \partial \vS^{-1} }{\partial \veta_C } \partial_{S^{-1}} \mathcal{L} \approx \frac{ \partial \vS^{-1} }{\partial \veta_C } \big[ \vg\vg^\top - \gamma (\vC_t\vC_t^\top)^{-1} \otimes (\vK \vK^\top)^{-1} \big ]= \sqrt{\frac{2}{d}} \vC^\top \vG^\top \vK \vK^\top  \vG \vC - \gamma \sqrt{2 d} \vI_p.
\end{align}

Thus, the update for block $\vC$ can be re-expressed as
\begin{align}
        \vC_{t+1} & \leftarrow \vC_t \mathrm{Exp}( - \frac{ \stepsize_2 }{2 d } \big( \vC^\top \vG^\top \vK \vK^\top  \vG \vC  -\gamma d \vI_p \big)  ) \approx \vC_t \big[  \vI - \frac{ \stepsize_2 }{2 d } \big( \vC^\top \vG^\top \vK \vK^\top  \vG \vC  -\gamma d \vI_p \big) \big]
\end{align}
We  also include a damping term  $\lambda \vI_{dp} = \lambda \vI_d \otimes \vI_p $ into the curvature approximation  such as $\mathcal{H}  = \vg\vg^\top + \lambda \vI_d \otimes \vI_p$. Recall that we do not assume that the curvature approximation $\mathcal{H}$  has the same structure as the preconditioner $\vS$.
We can similarly update blocks $\vK$ and $\vmu$. The deails of the complete update can be found at Fig.~\ref{fig:matrix_methos_full}.

\section{ Scale-invariance and Affine-invariance of Our Kronecker-factored Updates  }
\label{app:invariance_shampoo_root_free}

\begin{claim}
Our root-free Shampoo update shown in Fig.~\ref{fig:matrix_methos_simple} is invariant to a scale-transformation of the loss.
\end{claim}

\begin{proof}
We consider a matrix weight $\vmu=\mathrm{vec}(\vM)$. In this case, the (unscaled) optimization problem in \eqref{eq:optobj} is
\begin{align}
    \min_{M} \ell(\vM)
\end{align} where $\vM \in \real^{p \times d}$.

The scaled problem is
\begin{align}
 \min_{\bar{M}} \ell_{\text{scaled}}(\bar{\vM} ) \coloneqq  \frac{1}{B} \ell(\bar{\vM})
\end{align}

We will show that
our update is invariant at each step.
In other words, $\vM=\bar{\vM}$ holds for every iteration.
We assume $\vM=\bar{\vM}$ holds at the initialization step.

Recall that
our update rule  (Fig.~\ref{fig:matrix_methos_simple}) to solve the original problem is
\begin{align*}
    \vS_C &= (1-\stepsize_2 \gamma) \vS_C + \frac{\stepsize_2}{d} (\vG \vS_K^{-1} \vG^\top) \\
    \vS_K &= (1-\stepsize_2 \gamma) \vS_K + \frac{\stepsize_2}{p} (\vG^\top \vS_C^{-1} \vG) \\
    \vM & = \vM - \stepsize_1  \vS_C^{-1} \vG \vS_K^{-1}
\end{align*} where $\vG\coloneqq\nabla_M \ell(\vM) \in \real^{p \times d}$.

In this scaled case, our update shown in Fig.~\ref{fig:matrix_methos_simple} is
\begin{align*}
    \bar{\vS}_C &= (1-\stepsize_2 \gamma) \bar{\vS}_C + \frac{\stepsize_2}{d} ( {\color{red}B} \bar{\vG} \bar{\vS}_K^{-1} \bar{\vG}^\top) \\
    \bar{\vS}_K &= (1-\stepsize_2 \gamma) \bar{\vS}_K + \frac{\stepsize_2}{p} ({\color{red}B} \bar{\vG}^\top \bar{\vS}_C^{-1} \bar{\vG}) \\
    \bar{\vM} & = \bar{\vM} - \stepsize_1  \bar{\vS}_C^{-1} \bar{\vG} \bar{\vS}_K^{-1}
\end{align*} where
$\bar{\vG} =\nabla_{\bar{M}} \ell_{\text{scaled}}(\bar{\vM}) = \frac{1}{B} \vG $.

We first assume that  $\bar{\vS}_C = \frac{\vS_C}{\sqrt{B}} $ and $\bar{\vS}_K = \frac{\vS_K}{\sqrt{B}} $.
Therefore, we can  show that $\vM = \bar{\vM}$ since their descent directions are the same as
$\bar{\vS}_C^{-1} \bar{\vG} \bar{\vS}_K^{-1} = \big(\frac{\vS_C}{\sqrt{B}}\big)^{-1} \big(\frac{\vG}{B}\big) \big(\frac{\vS_K}{\sqrt{B}}\big)^{-1} = {\vS}_C^{-1} {\vG} {\vS}_K^{-1}$.

Now, we prove that  $\bar{\vS}_C = \frac{\vS_C}{\sqrt{B}} $ and $\bar{\vS}_K = \frac{\vS_K}{\sqrt{B}} $.
We will prove this by induction.
We assume that $\bar{\vS}_C $ and $\bar{\vS}_K$ are initialized so that
$\bar{\vS}_C = \frac{\vS_C}{\sqrt{B}} $ and $\bar{\vS}_K = \frac{\vS_K}{\sqrt{B}} $ hold in the base case needed for induction.
It is easy to see that
\begin{align*}
     \bar{\vS}_C^{\text{new}} &= (1-\stepsize_2 \gamma) \bar{\vS}_C + \frac{\stepsize_2}{d} ( {\color{red}B} \bar{\vG} \bar{\vS}_K^{-1} \bar{\vG}^\top)  \\
     &= (1-\stepsize_2 \gamma) \big(\frac{\vS_C}{\sqrt{B}}\big) + \frac{\stepsize_2}{d} ( B \frac{\vG}{B} \big( \frac{\vS_K}{\sqrt{B}} \big)^{-1}\frac{\vG}{B}^\top)  \\
     &= \frac{1}{\sqrt{B}}
     \big[ (1-\stepsize_2 \gamma) \vS_C + \frac{\stepsize_2}{d} (\vG \vS_K^{-1} \vG^\top) \big] \\
     &= \frac{\vS_C^{\text{new}}}{\sqrt{B}}
\end{align*} Thus, by induction, we can show
 $\bar{\vS}_C = \frac{\vS_C}{\sqrt{B}} $.
 We can similarly show that
 $\bar{\vS}_K = \frac{\vS_K}{\sqrt{B}}$.

\end{proof}

\begin{claim}
Our root-free Shampoo update shown in Fig.~\ref{fig:matrix_methos_simple} is affine reparametrization invariant up to an (Kronecker-factored) affine transformation.
\end{claim}

\begin{proof}
We consider the following unconstrained optimization problem with a matrix weight $\vM \in \real^{p \times d}$.
\begin{align}
    \min_{M} \ell(\vM)
\end{align}
Now, consider reparametrizing $\vM $ with known non-singular transformation matrices $\vE \in \real^{p \times p}$ and $\vF \in \real^{d \times d}$ and a constant matrix $\vK$ as $\vM = \vE \vN \vF^{T} +\vO$
In this case, the optimization problem becomes
\begin{align}
 \min_{ N } \ell^{\text{rep}}(\vN ) \coloneqq   \ell(\vE \vN \vF^\top +\vO)
\end{align}

Note that this is a Kronecker-factored affine transformation when we a vector representation.  In other words, $\mathrm{vec}(\vM) = (\vF \otimes \vE) \mathrm{vec}(\vN) + \mathrm{vec}(\vO)$, where
this transformation matrix $(\vF \otimes \vE)$ admits a Kronecker-factored structure.

We will show that our update is affine invariant for such a transformation at each step.
In other words, if we use the same update rule to solve these two problems, they are equivalent.

For the first problem,
 the method takes the following step at iteration $t$
\begin{align*}
  \vS_C^{(t+1)} &= (1-\stepsize_2 \gamma )\vS_C^{(t)} + \frac{\stepsize_2}{d} \vG^{(t)} \big( \vS_K^{(t)} \big)^{-1} \big( \vG^{(t)}\big)^\top,\\
   \vS_K^{(t+1)} &= (1-\stepsize_2 \gamma )\vS_K^{(t)} + \frac{\stepsize_2}{p} \big(\vG^{(t)}\big)^\top \big( \vS_C^{(t)} \big)^{-1}  \vG^{(t)},\\
 \vM^{(t+1)} &= \vM^{(t)} - \stepsize_1  \big( \vS_C^{(t+1)} \big)^{-1} \vG^{(t)} \big( \vS_K^{(t+1)}\big)^{-1}
\end{align*} where $\vG^{(t)} \coloneqq  \nabla_M \ell(\vM) \big|_{M= M^{(t)} } $.

For the second problem, we assume $\vS_{ C^{\text{rep}}}$, $\vS_{ K^{\text{rep}}}$, and $\vN$ are
initialized so that
$\vS_{ C^{\text{rep}}}^{(0)} = \vE^{-T} \vS_{ C}^{(0)}\vE^{-1} $,
$\vS_{ K^{\text{rep}}}^{(0)} = \vF^{-T} \vS_{ K}^{(0)}\vF^{-1} $,
and
$ \vE^{-1} (\vM^{(0)}-\vO) \vF^{-T} = \vN^{(0)} $.  %

In this case, the our update at the first iteration becomes
\begin{align*}
  \vS_{ C^{\text{rep}}}^{(t+1)} &= (1-\stepsize_2 \gamma )\vS_{ C^{\text{rep}}}^{(t)} + \frac{\stepsize_2}{d} \vG_{\text{rep}}^{(t)} \big( \vS_{ K^{\text{rep}}}^{(t)} \big)^{-1} \big( \vG_{\text{rep}}^{(t)}\big)^\top,\\
  \vS_{ K^{\text{rep}}}^{(t+1)} &= (1-\stepsize_2 \gamma )\vS_{ K^{\text{rep}}}^{(t)} + \frac{\stepsize_2}{p} \big( \vG_{\text{rep}}^{(t)}\big)^\top \big( \vS_{ C^{\text{rep}}}^{(t)} \big)^{-1} \vG_{\text{rep}}^{(t)},\\
 \vN^{(t+1)} &= \vN^{(t)} - \stepsize_1  \big( \vS_{ C^{\text{rep}}}^{(t+1)} \big)^{-1} \vG_{\text{rep}}^{(t)} \big( \vS_{ K^{\text{rep}}}^{(t+1)}\big)^{-1}
\end{align*} where we use the following identities when $t=0$.
\begin{align}
    \vG_{\text{rep}}^{(t)} & \coloneqq  \nabla_N \ell^{\text{rep}}(\vN) \big|_{N= N^{(t)} } = \frac{\partial \vM}{\partial \vN}|_{M= M^{(t)} } \nabla_M \ell^{\text{rep}}(\vM) \big|_{M= M^{(t)} } =  \vE^\top \nabla_M \ell(\vM)|_{M= M^{(t)} } \vF = \vE^\top \vG^{(t)}  \vF
\end{align}
It is easy to see that the following expressions hold when $t=0$.
\begin{align*}
 \vE^{-T} \vS_{ C^{\text{rep}}}^{(t+1)} \vE^{-1} &= (1-\stepsize_2 \gamma ) \vE^{-T} \vS_{ C^{\text{rep}}}^{(t)} \vE^{-1} + \frac{\stepsize_2}{d} \vE^{-T} \underbrace{\vG_{\text{rep}}^{(t)}}_{E^\top G^{(t)} F} \big(  \vS_{ K^{\text{rep}}}^{(t)}   \big)^{-1} \big( \vG_{\text{rep}}^{(t)}\big)^\top \vE^{-1},\\
 &= (1-\stepsize_2 \gamma ) \underbrace{ \vE^{-T} \vS_{ C^{\text{rep}}}^{(t)} \vE^{-1} }_{  S_{ C}^{(t)} } + \frac{\stepsize_2}{d} \vG^{(t)} \big( \underbrace{ \vF^{-T} \vS_{ K^{\text{rep}}}^{(t)} \vF^{-1} }_{ S_{K}^{(t)} } \big)^{-1}  \big( \vG^{(t)}\big)^\top \\
 &= (1-\stepsize_2 \gamma )  \vS_{ C}^{(t)} + \frac{\stepsize_2}{d} \vG^{(t)} \big(  \vS_{K}^{(t)}  \big)^{-1}  \big( \vG^{(t)}\big)^\top  \\
 &=
 \vS_{C}^{(t+1)}
\end{align*}
Similarly, we can show
$\vF^{-T} \vS_{ K^{\text{rep}}}^{(t+1)} \vF^{-1}= \vS_{K}^{(t+1)}$ when $t=0$.

Consequently, we have the following result when $t=0$
\begin{align*}
 \vE \vN^{(t+1)} \vF^\top &=  \vE\vN^{(t)}  \vF^\top - \stepsize_1 \vE  \big( \vS_{ C^{\text{rep}}}^{(t+1)} \big)^{-1} \underbrace{ \vG_{\text{rep}}^{(t)} }_{E^\top G^{(t)}F} \big( \vS_{ K^{\text{rep}}}^{(t+1)}\big)^{-1} \vF^\top \\
 &= \underbrace{\vE\vN^{(t)}  \vF^\top}_{ M^{(t)} - O } - \stepsize_1 \underbrace{ \vE  \big( \vS_{ C^{\text{rep}}}^{(t+1)} \big)^{-1} \vE^\top}_{ \big(S_C^{(t+1)}\big)^{-1} } \vG^{(t)} \underbrace{ \vF \big( \vS_{ K^{\text{rep}}}^{(t+1)}\big)^{-1} \vF^\top}_{ \big(S_K^{(t+1)}\big)^{-1}  } = \vM^{(t+1)} -\vO.
\end{align*}
In other words, $\vM^{(t+1)}= \vE \vN^{(t+1)} \vF^\top + \vO$ when $t=0$.
From this  expression, we can see that both updates are equivalent at the first iteration.
Similarly, we can show that both updates are equivalent at every iteration by induction.

\end{proof}

We can also show that our inverse-free Shampoo update shown in Fig.~\ref{fig:matrix_methos_simple} is both scale and affine invariant.

\section{Preconditioner Invariance of Our Kornecker-factored Updates }
\label{app:invariance_ours_root_free}

We reparametrize a Kronecker-factored  preconditioner $\vS=\vS_C \otimes \vS_K$ to obtain an inverse-free update scheme.
The following claim explicitly demonstrates the preconditioner invariance.
In other words, the updates of $\vC$ and $\vS_C$ are equivalent up to  first-order accuracy.

\begin{claim}
\label{claim:shampoo_invariance_free}
Let  $\vC\vC^\top $ be the inverse of a preconditioner  updated according to the inverse-free scheme
in Fig.~\ref{fig:matrix_methos_simple}
 with initialization ${\vC\vC} = \vS_C^{-1}$.
 If  $\vC$ and $\vS_C$ are updated by using the same sequence of gradients $\vG$, then
   $\vC\vC^\top$ has  first-order accuracy of our root-free Shampoo update of $\vS_C$
in Fig.~\ref{fig:matrix_methos_simple} at each iteration, i.e., \( \vC\vC^\top  =    \vS_C^{-1}  + O(\stepsize_2^2)\).
Similarly,  $\vK\vK^\top$ has  first-order  accuracy of our root-free update of $\vS_K$ at each iteration, i.e., \( \vK\vK^\top  =    \vS_K^{-1}  + O(\stepsize_2^2)\).
\end{claim}

\begin{proof}
It is equivalent to show that  $ (\vC\vC^\top) \vS_C =   \vI   + O(\stepsize_2^2)$ and
 $ (\vK\vK^\top) \vS_K =   \vI   + O(\stepsize_2^2)$.
We will prove this  by induction. Thanks to the initialization, we know that the base case ($t=0$) is true.
Now, we assume this relationship holds when $t=k$.
Consider the case when $t=k+1$.

For notation simplicity, we drop the index $k$ and
denote
$\vC^{(k+1)}$ and $\vS_C^{(k+1)}$ by $\vC^{\text{new}}$ and $\vS_C^{\text{new}}$.
We have the following result
\begin{align*}
     &\big[\vC^{\text{new}}\big(\vC^{\text{new}}\big)^\top\big] \vS_C^{\text{new}} \\
      = & \vC \big[ \vI - \frac{\stepsize_2}{d} \big(
     \vC^\top \vG \vK\vK^\top \vG^\top \vC - \gamma d \vI \big)
    + O(\stepsize_2^2)  \big] \vC^\top \big[ (1- \gamma\stepsize_2 ) \vS_C + \frac{\stepsize_2}{d} \vG \vS_K^{-1}\vG^\top \big] \\
     =& \big[ (1+ \gamma \stepsize_2) \vC\vC^\top - \frac{\stepsize_2}{d}  \vC \vC^\top \vG \vK\vK^\top \vG^\top \vC\vC^\top \big] \big[ (1- \gamma\stepsize_2 ) \vS_C + \frac{\stepsize_2}{d}\vG \vS_K^{-1}\vG^\top \big]+ O(\stepsize_2^2)   \\
     =& \underbrace{ \vC\vC^\top \vS_C}_{ I + O(\stepsize_2^2) } - \frac{\stepsize_2}{d} \vC \vC^\top \vG \vK\vK^\top \vG^\top \underbrace{\vC\vC^\top\vS_C}_{I + O(\stepsize_2^2)} +  \frac{\stepsize_2}{d} \vC\vC^\top \vG \underbrace{\vS_K^{-1} }_{KK^\top + O(\stepsize_2^2)}\vG^\top + O(\stepsize_2^2) \\
     =& \vI + O(\stepsize_2^2)
\end{align*}
Likewise, we have
$\big[\vK^{\text{new}}\big(\vK^{\text{new}}\big)^\top\big] \vS_K^{\text{new}} = \vI + O(\stepsize_2^2)$.
Thus, we can use induction to show the relationship holds at every iteration.
\end{proof}

As discussed in the caption of Fig.\ref{fig:matrix_methos_full}, we can make the higher order term $O(\stepsize_2^2)$ negligible when $\stepsize_2$ is determined by the matrix norm of
$\frac{1}{d} \big(
     \vC^\top \vG \vK\vK^\top \vG^\top \vC - \gamma d \vI \big)$. %

\section{ Experimental Details \& Additional Experiments  }
\label{sec:app_exp_nn}

We consider various NN models ranging from classical to modern models to demonstrate the effectiveness of root-free adaptive methods. We consider the following NN models in our (pre-)training experiments.
\begin{itemize}
    \item CNNs: ResNet34, VGG16, DenseNet121 on CIFAR100
    \item RNN: 3-layer LSTM on PenTree
    \item GNN:  Graph MLP with attention on OgbnProducts
    \item Transformers: SwinViT, FocalNet, GCViT on ImageWoof10
    \item Mamba: VMamba on ImageWoof10
\end{itemize}
We train CNNs for 210 epochs, and ViTs and VMamba for 300 epochs with mini-batch size 128. For CNNs, RNN, and GNN, we use a step decay schedule suggested by \citet{wilson2017marginal}.
For Transformers and Mamba, we use a cosine learning rate schedule suggested by \citet{chen2023symbolic}.
Our implementation can be found at \url{https://github.com/yorkerlin/remove-the-square-root}.

For fine-tuning experiments, we use pre-trained weights of a ViT model
 \citep{dosovitskiy2020image}  and fine-tune them on CIFAR100 and FOOD101 datasets. We use this implementation (\url{https://github.com/bwconrad/vit-finetune}) in our experiments.

\paragraph{Hyperparameter Tuning}
We use PyTorch's built-in SGD, AdamW, and RMSProp. For Shampoo, we rely on the state-of-the-art PyTorch implementation from Meta \citep{shi2023distributed}.
We tune the following hyperparameters (HPs) for each optimizer.
\begin{itemize}
    \item SGD: initial learning rate, momentum, weight decay
    \item AdamW: initial learning rate, coefficients used for estimating the first and second moment, weight decay, damping
    \item RMSProp: initial learning rate, coefficient used for estimating the second moment,  momentum, weight decay, damping
    \item Shampoo: initial learning rate, coefficients used for updating momentum and Kronecker factors, weight decay, damping
    \item RF-RMSProp (ours, c.f. Fig.~\ref{fig:rmsprop-full}): initial learning rate, coefficients used for estimating the second moment,  momentum, weight decay, damping
    \item IF-Shampoo with $\gamma=1$ (ours, c.f.  Fig.~\ref{fig:matrix_methos_full}): initial learning rate, coefficients used for updating momentum and Kronecker factors, weight decay, damping, Riemannian momentum
\end{itemize}

For matrix adaptive methods (Shampoo and IF-Shampoo), we update their matrix preconditioners at each 2 iterations.
For all tasks and optimizers, we employ a two-stage HP tuning protocol based on random search \citep{choi2019empirical}.
Unlike \citet{choi2019empirical}, we only consider a small damping term (e.g., $0<\lambda<10^{-4}$) in our HP search for all these methods.
In the first stage, we use larger search regimes for all HPs. Based on this stage, we select a narrower HP range and re-run the search, reporting the best run for each method.  In each stage, we use 100 runs. We use this implementation (\url{https://github.com/yorkerlin/remove-the-square-root}) to conduct our HP search in the second stage.

\paragraph{Mixed-precision Training}
For all optimizers, only the forward pass is executed in mixed precision with BFP-16 (as recommended by the official PyTorch guide). The gradients are automatically cast back to FP-32 by PyTorch. Shampoo uses these FP-32 gradients for its preconditioner and is unstable when converting them to BFP-16 \citep{shi2023distributed} . Instead, our IF-Shampoo converts the gradients into BFP-16, updates the preconditioner, and even takes preconditioned gradient steps (including momentum) in half precision. Our method works well in half precision without using matrix decomposition and matrix solve/inversion. These matrix operations in half precision are not supported in PyTorch and JAX because they are numerically unstable.

\begin{figure}[H]
  \centering
  \includegraphics[width=\linewidth]{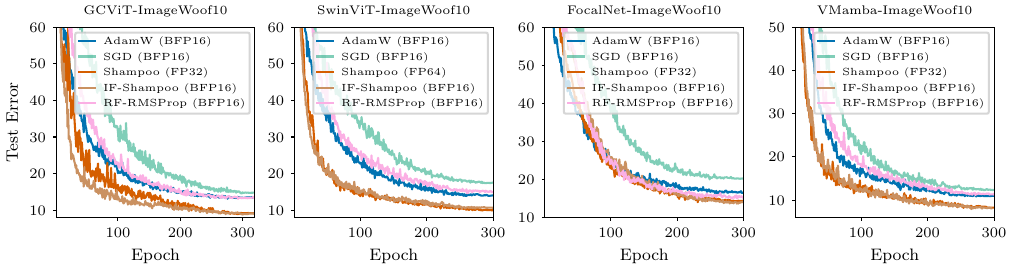}
  \vspace{-6ex}
  \caption{
  Comparison of matrix root-free versus root-based methods on GCViT~\citep{hatamizadeh2023global}, SwinViT~\citep{liu2021swin},  FocalNet~\citep{yang2022focal}, and VMamba~\citep{liu2024vmamba}.
  Both matrix methods (Shampoo, IF-Shampoo) outperform diagonal  methods on modern vision models such as transformers and mambas using modern training strategies  (cosine learning rate schedule, random search using 200 runs).
  In contrast to Shampoo, our inverse-free matrix  method, IF-Shampoo, runs in BFP-16 and trains at least twice as fast, while using less memory. We update
  matrix preconditioners  at each 2 iterations and
  can further reduce the wall clock time by updating them less frequently.
\vspace{-0.44cm}
  }
  \label{fig:experiment-matrix-more}
\end{figure}

\begin{table}[H]
\begin{minipage}[t]{\linewidth}
\centering
 \begin{tabular}{c c c c c c}
		\toprule
		\begin{tabular}{c}
                  Dataset  \\
		\end{tabular} &
		Method &
		\begin{tabular}{c}
			  GCViT  \\
		\end{tabular} &
		\begin{tabular}{c}
                  SwinViT  \\
		\end{tabular} &
		\begin{tabular}{c}
                FocalNet  \\
		\end{tabular}&
  		\begin{tabular}{c}
			 VMamba   \\
		\end{tabular}
		\\
                \midrule
		\multirow{3}{*}{
		  \begin{tabular}{c}
                    \\
                 ImageWoof10
		\end{tabular}}
                & AdamW        & $13.48/143$    & $13.96/162$    & $16.58/159$   &  $10.96/315$    \\
		& SGD   & $14.74/142$  & $17.44/161$     & $20.22/158$    & $12.38/314$    \\
		& Shampoo      & $9.16/550$   &  $9.97/633$ &  $14.33/716$ &  $8.25/1336$ \\
		& RF-RMSProp (ours)       & $13.37/144$   &  $15.02/162$ &  $15.39/159$ &  $11.38/315$ \\
        & IF-Shampoo (ours)  & $9.05/202$ &  $10.65/169$ &  $13.93/189$  & $8.20/384$ \\
                \bottomrule \\
	\end{tabular}
\caption{
Results about the performance (test error/wall clock time) of the  optimizers on modern NN models. We train all models for 300  epochs.
All methods except Shampoo support training with BFP-16.
Shampoo has to use FP-32---sometimes FP-64---to update its preconditioners to avoid numerical instabilities. For example, Shampoo has to use FP-64 on the SwinViT model.
For matrix adaptive methods (Shampoo, IF-Shampoo), we update their preconditioners at every 2 iterations and can further reduce the wall clock time by updating them less frequently.
The results are obtained by averaging over the last 10 iterations.   }
\label{table:results2}
\end{minipage}
\end{table}

\begin{figure}[H]
  \centering
  \includegraphics[width=\linewidth]{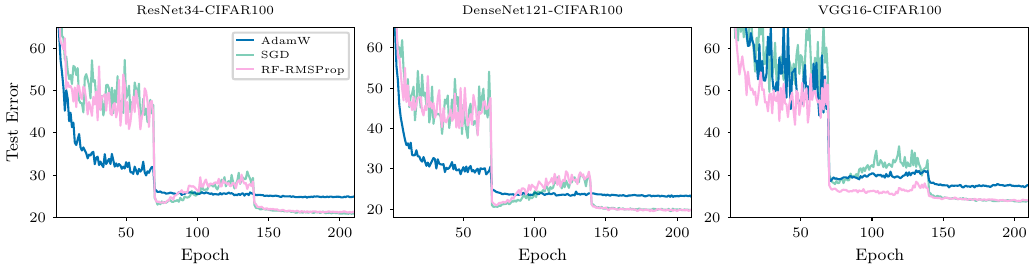}
  \vspace{-6ex}
  \caption{
  Comparison of root-free versus root-based adaptive methods on ResNet34, DenseNet121, and VGG16 using modern training strategies  (step decay learning rate schedule, random search using 200 runs).
  Root-free adaptive methods close the generalization gap between their root-based counterparts and SGD on CNNs.
\vspace{-0.44cm}
  }
  \label{fig:experiment-cnn-more}
\end{figure}

\begin{table*}[h]
\begin{minipage}[t]{\linewidth}
\centering
 \begin{tabular}{c c c c c c}
		\toprule
		\begin{tabular}{c}
                  Dataset  \\
		\end{tabular} &
		Method &
		\begin{tabular}{c}
                  ResNet34  \\
		\end{tabular} &
		\begin{tabular}{c}
			  DenseNet121  \\
		\end{tabular} &
		\begin{tabular}{c}
			  VGG16  \\
		\end{tabular}
		\\
                \midrule
		\multirow{3}{*}{
		  \begin{tabular}{c}
                  CIFAR100
		\end{tabular}}
                & AdamW        & $24.85$     &  $23.30$ & $27.42$    \\
		& SGD   & $20.91$  & $19.88$    & $24.02$   \\
		& RF-RMSProp (ours)   & $21.25$  &  $19.81$  & $24.00$\\
                \bottomrule \\
	\end{tabular}
\caption{
Results about the performance (test error) of the  optimizers on convolutional  NN models.
The results are obtained by averaging over the last 10 iterations.   }
\label{table:results3}
\end{minipage}
\end{table*}

\begin{figure}[H]
  \centering
  \includegraphics[width=\linewidth]{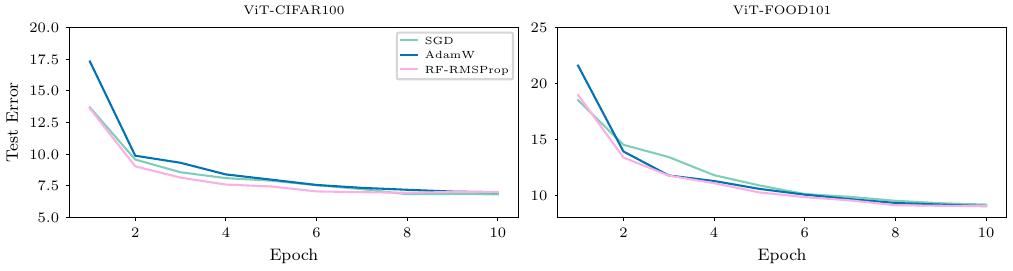}
  \vspace{-6ex}
  \caption{Experiments demonstrating that square-root-free adaptive methods work well in fine-tuning settings, where we use a ViT model \citep{dosovitskiy2020image} pre-trained on ImageNet-21k and fine-tune it on  CIFAR100 and FOOD101 datasets.
\vspace{-0.42cm}
  } \label{fig:experiment-fine_tune}
\end{figure}

\paragraph{Additional Experiments}
We also evaluate our root-free methods on fine-tuning problems (c.f. \ref{fig:experiment-fine_tune}).

\end{document}